\documentclass[letterpaper, 10 pt, conference]{ieeeconf}  

\IEEEoverridecommandlockouts                              

\usepackage{graphics} 
\usepackage{epsfig} 
\usepackage{url}
\usepackage{subfigure}
\usepackage{xcolor}
\usepackage{amsmath} 
\usepackage{amssymb}  
\usepackage[linesnumbered,ruled,vlined]{algorithm2e}
\usepackage{booktabs}
\usepackage{multirow}
\usepackage{cite}
\newtheorem{defn}{Definition}

\newcommand{\tool}{F\textsc{lyover}\xspace}
\newcommand{\website}{\url{https://ntutangyun.github.io/highway-interchange-dataset-website/}\xspace}

\title{\LARGE \bf
\tool: A Model-Driven Method to Generate Diverse Highway Interchanges for Autonomous Vehicle Testing
}

\author{Yuan Zhou$^{1*}$, 
Gengjie Lin$^{2*}$, 
Yun Tang$^{3}$, 
Kairui Yang$^{2}$, 
Wei Jing$^{2}$, 
Ping Zhang$^{2\ddagger}$, \\ 
Junbo Chen$^{2}$, 
Liang Gong$^{4}$, and 
Yang Liu$^{1}$ 
\thanks{$^*$ Equal contribution; $^\ddagger$ Corresponding author.}
\thanks{$^{1}$Yuan Zhou and Yang Liu are with the School of Computer Science and Engineering, Nanyang Technological University, Singapore
    {\tt\small \{y.zhou, yangliu\}@ntu.edu.sg}}
\thanks{$^{2}$Gengjie Lin, Kairui Yang, Wei Jing, Ping Zhang and Junbo Chen are with DAMO Academy, Alibaba Group, China.
    {\tt\small lingengjie@sjtu.edu.cn, \{kairui.ykr, jw334405\}@alibaba-inc.com,  junbo.chenjb@taobao.com, zp.zp@cainiao.com}}
\thanks{$^{3}$Yun Tang is with Alibaba-NTU Singapore Joint Research Institute, Nanyang Technological University, Singapore.
    {\tt\small yun005@e.ntu.edu.sg}}
\thanks{$^{4}$Liang Gong is with School of Mechanical Engineering, Shanghai Jiao Tong University, China.
    {\tt\small gongliang\_mi@sjtu.edu.cn}}
}

\begin{document}

\maketitle
\thispagestyle{empty}
\pagestyle{empty}

\begin{abstract}
It has become a consensus that autonomous vehicles (AVs) will first be widely deployed on highways.
However, the complexity of highway interchanges becomes the bottleneck for deploying AVs.
An AV should be sufficiently tested  under different highway interchanges, which is still challenging due to the lack of available datasets containing diverse highway interchanges.
In this paper, we propose a model-driven method, \tool, to generate a dataset consisting of diverse interchanges with measurable diversity coverage.
First, \tool proposes a labeled digraph to model the topology of an interchange.
Second, \tool takes real-world interchanges as input to guarantee topology practicality and extracts different topology equivalence classes by classifying the corresponding topology models.
Third, for each topology class, \tool identifies the corresponding geometrical features for the ramps and generates concrete interchanges using k-way combinatorial coverage and differential evolution.
To illustrate the diversity and applicability of the generated interchange dataset, we test the built-in traffic flow control algorithm in SUMO and the fuel-optimization trajectory tracking algorithm deployed to Alibaba's autonomous trucks on the dataset.
The results show that except for the geometrical difference, the interchanges are diverse in throughput and fuel consumption under the traffic flow control and trajectory tracking algorithms, respectively.

\end{abstract}


\section{Introduction}

Autonomous vehicles (AVs), such as autonomous sedan cars and autonomous trucks, play an essential role in relieving traffic congestion and eliminating accidents in future intelligent transportation systems.
AVs will first be deployed on highways since, compared with their urban counterparts, highway transport systems show less complexity.
The complexity of highways mainly lies in highway interchanges due to the diverse connection topology and ramp geometry, which significantly affect the decision-making and motion control of AVs.
Therefore, the reliability guarantee in highway interchanges becomes the bottleneck of their wide deployment in highway transport systems~\cite{zhu2022merging,sun2017capacity}.
To guarantee safety, researchers have designed different planning and control algorithms to ensure AVs move across interchanges successfully and smoothly~\cite{evaluation_merge,evaluation_speed,evaluation_motion,wang2021intelligent,claussmann2019review}.
However, these methods are tested under only some specific interchanges and lack a comprehensive evaluation for different interchanges.

Comprehensive AV testing under diverse interchanges is challenging.
On the one hand, it is risky, resource-consuming, and even impossible to cover all interchanges in a city physically.
On the other hand, even though simulation-based testing provides an efficient way to perform AV testing, it highly relies on HD maps. The lack of a ready-to-use dataset consisting of diverse highway intersections becomes the bottleneck.

There is little work on the generation of diverse highway interchanges.
In \cite{generate_lidar}, the authors propose a learning-based method to construct lane-level HD maps by fusing LiDAR point cloud and camera image features.
The authors in \cite{generate_gps} present an approach to generating a road network map from GPS trajectories, where the   GPS data is first processed to construct a road bitmap, and then the skeleton on the bitmap is computed.
This network map contains only the topology of the road without the height information.
In \cite{generate_virtual}, a real-time model is introduced to simulate the LiDAR sensor for an automotive, where the model can generate 3D point cloud maps. However, it is difficult to simulate real driving behavior in the virtual scenario.
Moreover, all the above methods aim to generate real-world maps, which cannot measure and guarantee the diversity of highway interchanges.

Hence, in this paper, we propose a model-driven method, \tool, to generate diverse highway interchanges.
First, we use labeled digraphs to model the topology of highway interchanges, where the highway roads and ramps are modeled as nodes, their connections are modeled as directed arcs, and the labels associated with the arcs identify the relative directions of the two nodes.
Second, we retrieve real-world interchanges from some map application to generate a set of practical topologies and extract the corresponding topology models.
Based on the topology models, we classify the topologies into equivalent classes.
For each topology class, we use k-way combinatorial coverage to sample a set of geometrical features and use differential evolution to generate concrete interchanges for each geometrical feature.
Finally, we test the throughput of the built-in traffic flow control algorithm in SUMO and the fuel consumption of the trajectory tracking deployed to Alibaba's autonomous trucks on the generated dataset.
The results show that the generated dataset consists of diverse interchanges and can be applied to test AV algorithms.

The contributions are as follows:
\begin{itemize}
\item We propose a method to model the highway interchange topology using a labeled digraph.
\item We propose a real-world interchange-induced method to generate practical interchange topologies.
\item We propose a coverage-guided method to generate concrete highway interchanges.
\item We build the first highway interchange dataset for simulation-based AV testing.
\end{itemize}

\section{Background and Problem Statement}

\begin{figure}
    \centering
    \subfigure[Interchange $J_1$ diagram]{
        \begin{minipage}[b]{0.47\columnwidth}
            \centering
            \label{fig:example_interchange_diagram}
            \includegraphics[width=1\textwidth]{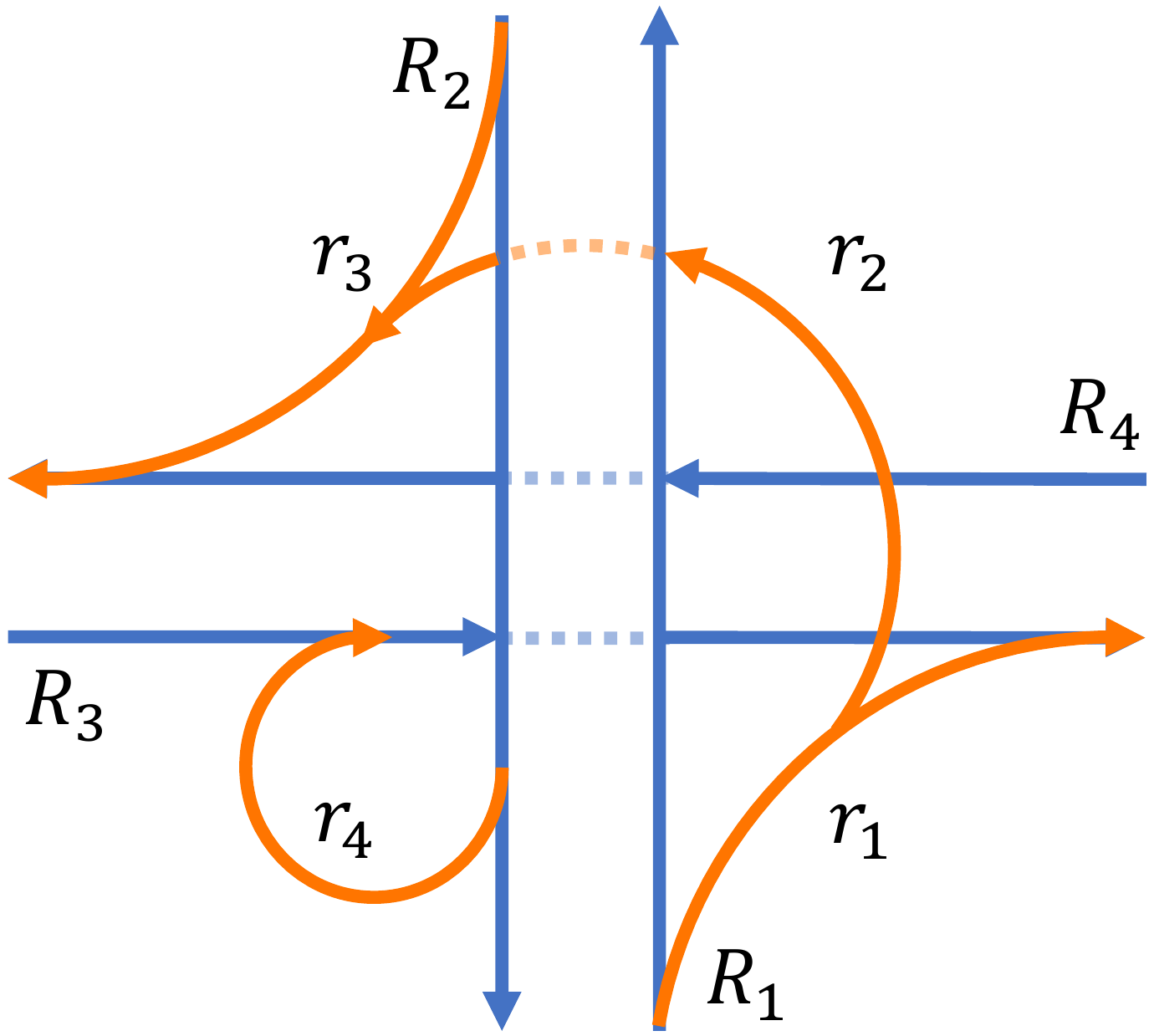}
        \end{minipage}}
    \subfigure[Topology graph of $J_1$]{
        \begin{minipage}[b]{0.47\columnwidth}
            \centering
            \label{fig:example_interchange_graph}
            \includegraphics[width=1\textwidth]{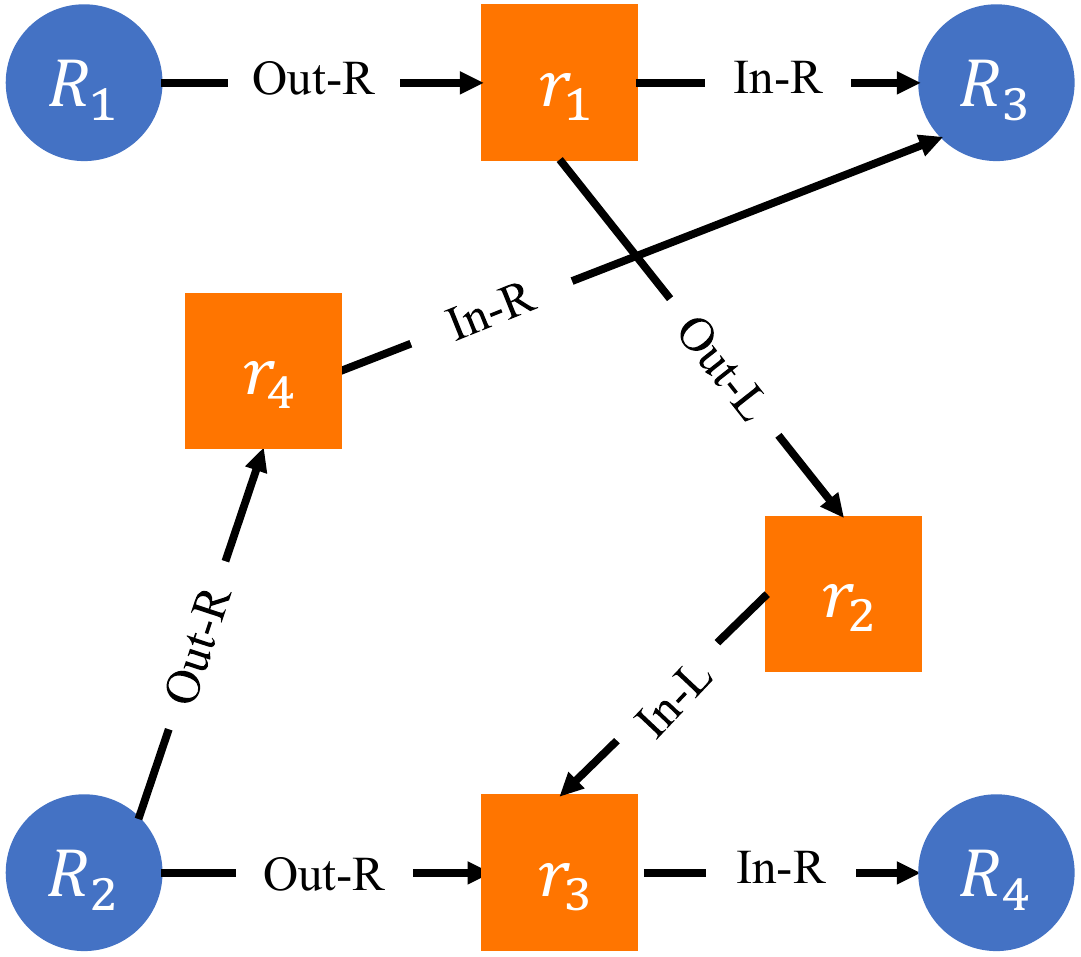}
        \end{minipage}}
    \caption{Example interchange and its corresponding topology graph}
    \label{fig:interchange_diagram_graph}
\end{figure}

A \textit{highway interchange} is usually formed at the intersection by multiple expressways.
Each expressway may contain either a single \textit{one-way road} or two parallel one-way roads in the opposite direction.
\textit{Ramps} are constructed to connect different one-way roads.
For example, as shown in Fig.~\ref{fig:example_interchange_diagram}, $J_1$ s formed by two crossing expressways, each containing two one-way roads (i.e., $\{R_1, R_2\}$ and $\{R_3, R_4\}$, respectively), and four ramps (i.e., $r_1-r_4$).


With different landforms, highway interchanges in different regions show various topologies and geometries.
On the one hand, the topology of an interchange is characterized by the number of one-way roads and ramps and their connections.
First, different interchanges may contain a different number of one-way roads. For example, there are four one-way roads in $J_1$, and a more complex interchange may have more one-way roads.
Second, one-way roads can be connected by different ramps.
For example, in $J_1$, the road pair $(R_2, R_3)$ is connected by ramp $r_4$, while $(R_1, R_4)$ is connected by the ramps $r_1$, $r_2$, and $r_3$.
Moreover, a ramp can leave or enter a road or another ramp from the left or right.
Therefore, different topologies can generate different paths using the planning algorithms, resulting in different traffic throughputs.
On the other hand, the geometrical characteristics, such as the number of lanes in each one-way road and the curvature and slope of each ramp, will affect the computation of control efforts.
First, the number of lanes in a one-way road can affect an AV's decision-making, which further affects the traffic throughput in the interchange as performing more lane changes will block other vehicles.
Second, each ramp has a speed limit according to its maximal curvature to guarantee safety.
Therefore, a vehicle must decelerate or accelerate to match the speed limit before entering a ramp. In addition, different slopes affect the control efforts, such as the throttle and brake.
Hence, the geometrical and topological profiles essentially affect the performance of AVs.

However, given the intersecting expressways, we can design unlimited types of interchanges with diverse geometrical and topological profiles.
So, how to measure the diversity of an interchange and how to generate a finite set of interchanges that can guarantee diversity should be addressed urgently.
Hence, our problem can be stated as follows:

\noindent \textbf{Problem}:
\textit{Propose a systematic method to generate diverse interchanges with measurable diversity coverage.}

\section{Methodology}

\begin{figure}
    \centering
    \includegraphics[width=\columnwidth]{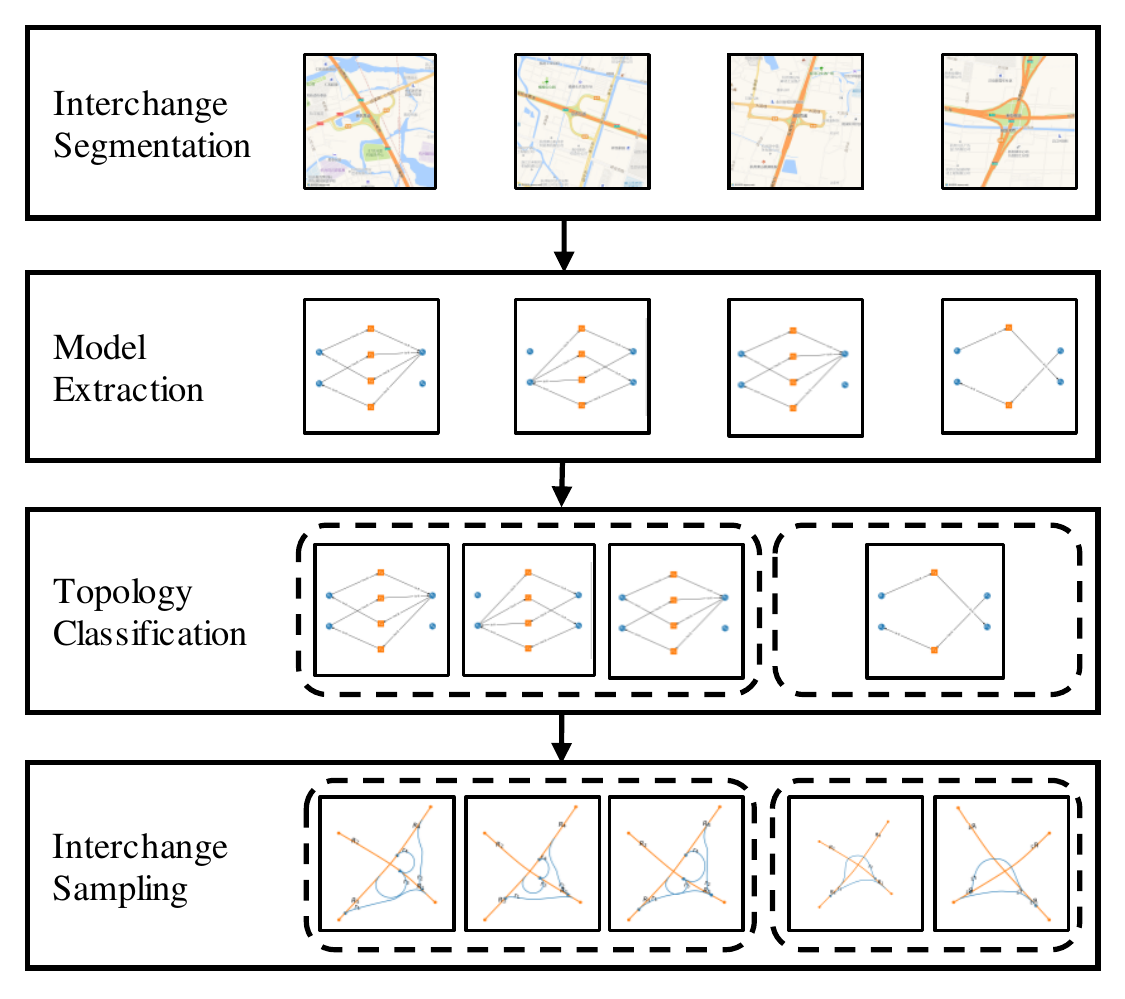}
    \vspace{-15pt}
    \caption{Overview of \tool.}
    \label{fig:method_overview}
\end{figure}

To address the problem, we propose a model-based method, \tool, to generate diverse interchanges with the guarantee of topology feasibility in the real world.
The overview of \tool is presented in Fig.~\ref{fig:method_overview}.
\tool first extracts the 2D real-world interchange maps from an existing map application, such as GaoDe Map.
For each interchange, \tool extracts the roads, ramps, and their connectivity relationships and models its topology with labeled digraph.
Third, \tool classifies the topology models into topology-equivalent classes in terms of graph isomorphism.
Finally, for each topology class, \tool generates concrete highway interchanges using k-way combinatorial coverage and differential evolution.

\subsection{Modeling of Interchange Topology}

We apply labeled digraph \cite{von1988pattern} to model interchange typologies.

\begin{defn}
A labeled digraph $\mathbb{G}$ is a tuple $\langle V, E, T,  f_E\rangle$, where
\begin{itemize}
    \item $V$ is a finite set of labeled vertices;
    \item $E\subseteq V\times V$ is a set of directed edges;
    \item $T$ is a set of labels;
    \item $f_E: E\to T$ is a label mapping function assigning each edge a label;
\end{itemize}
\end{defn}

Given an interchange $J$ with $n$ one-way roads and $m$ ramps, its topology model, denoted as $\mathbb{G}^J= \langle V, E, T, f \rangle$, can be constructed as follows.
(1) Suppose the sets of one-way roads and ramps are $V^{road}=\{R_i, i\in\mathbb{N}_n\}$ and $V^{ramp}=\{r_k, k\in \mathbb{N}_m\}$, where $\mathbb{N}_n=\{1,2,\ldots, n\}$.
Hence, $V=V^{road} \cup V^{ramp}$.
(2) Given $v_1, v_2\in V$, $(v_1, v_2)\in E$ if $v_1$ and $v_2$ are connected and adjacent, i.e., a vehicle can change lanes from $v_1$ to $v_2$ directly.
(3) The labels denote how a ramp connects to a road or another ramp.
Hence, we have four labels: $T=\{$Out-R, Out-L, In-R, In-L$\}$, where Out-R (resp., Out-L) means the ramp leaves a road$/$ramp on its right (resp., left), and In-R (resp., In-L) means the ramp merges into a road$/$ramp from its right (resp., left).
Therefore, given an edge $(v_1, v_2)$, the label mapping function can be defined as:
\begin{equation*}
    f(v_1,v_2) = \left\{
\begin{aligned}
    \text{Out-R} & \text{ if $v_2$ leaves $v_1$ on its right;}\\
    \text{Out-L} &\text{ if $v_2$ leaves $v_1$ on its left;}\\
    \text{In-R}  &\text{ if $v_2$ merges to $v_1$ from the right side;}\\
    \text{In-L} & \text{ if $v_2$ merges to $v_1$ from the  left side;}
\end{aligned}
    \right.
\end{equation*}

For example, Fig. \ref{fig:example_interchange_graph} presents the topology model of the interchange $J_1$ (Fig.~\ref{fig:example_interchange_diagram}).
We have $\mathbb{G}^{J_1} = \langle V_1, E_1, T_1, f_1 \rangle$, where $V^{road}_1=\{R_1, R_2, R_3, R_4\}$, $V^{ramp}_1=\{r_1, r_2, r_3, r_4\}$, $E_1 = \{(R_1, r_1)$, $(r_1, R_3)$, $(r_1, r_2)$, $(r_2, r_3)$, $(R_2, r_3)$, $(r_3, R_4)$, $(R_2, r_4)$, $(r_4, R_3)\}$ and
$f_1(R_1, r_1) = f_1(R_2, r_3)= f_1(R_2, r_4) = \text{Out-R}$,
$f_1(r_1,r_2) = \text{Out-L}$,
$f_1(r_1,R_3) = f_1(r_4, R_3)= f_1(r_3, R_4) = \text{In-R}$,
$f_1(r_2,r_3) = \text{In-L}$.

\subsection{Topology-Based Interchange Classification}
Clearly, the topology model describes the connection pattern of an interchange.
Interchanges with different topology models show different road connections and thus affect vehicles' motion behaviors.
Hence, in this subsection, we classify interchanges based on their topology models.

According to the isomorphism of labeled graphs~\cite{hsieh2006efficient}, we first define the topology isomorphism of two interchanges.
\begin{defn}
Given two interchanges $J$ and $J'$, and their topology models $\mathbb{G}=\langle V, E, T, f_E \rangle$ and $\mathbb{G}'=\langle V', E', T', f'_E \rangle$, $\mathbb{G}$ and $\mathbb{G}'$ are isomorphic, denoted as $\mathbb{G} \cong \mathbb{G}'$, if there exists a bijective function $g: V \to V'$ such that:\\
(1) $\forall u,v\in V$, $(u,v) \in E \Leftrightarrow (g'(u), g'(v))\in E'$, and\\
(2) $\forall (u,v)\in E$, $f_E(u,v) = f'_E(g(u), g(v))$.\\
$J$ and $J'$ are called topology isomorphism.
\end{defn}


\begin{defn}\label{def:topology-classification}
Two interchanges $J_1$ and $J_2$ are topology equivalent if they are topology isomorphism.
\end{defn}


Based on Definition~\ref{def:topology-classification}, we can classify interchanges into different equivalency classes.
The main process is to check whether two topology models are isomorphic, which is resolved by  VF2~\cite{cordella2001improved} in \tool.

\subsection{Coverage-Guided Interchange Sampling}
The topology modeling and classification process quantifies the interchange diversity at the topological level. In this subsection, we further explore the diversity at the geometric level when generating concrete interchanges.

To generate concrete interchanges from each topology class, we need to define the geometries of the roads and ramps.
Usually, there are many geometrical features to describe a road$/$ramp, such as road length, lane width, and lane curvature.
In this paper, we mainly focus on the planning and control modules of an AV, so we identify the following features:
1) \textit{the number of lanes} for a one-way road, which affects the local trajectory computation of an AV,
2) \textit{minimum radius} of a ramp, which determines the speed limit of the ramp, and
3) \textit{maximum longitudinal slope} of a ramp, which has a significant impact on the throttle control of an AV.
Hence, the feature of an interchange class can be defined as:
\begin{defn}
Given an interchange class $J$, the \emph{interchange feature} is a tuple $F^J_i =\langle \mathbb{G}^J, F_i(R^J), F_i(r^J) \rangle$, where
\begin{itemize}
    \item $\mathbb{G}^J=\langle V^{road}\cup V^{ramp}, E, T, f_E \rangle$ is the topology model;
    \item $F_i(R^J)=\{L_i(v): v\in V^{road}\}$, and $L_i(v)\in \mathbb{N}$ denotes the number of lanes in road $v$, where $\mathbb{N}$ is the set of natural numbers;
    \item $F_i(r^J)=\{(c_i(v), l_i(v)): v\in V^{ramp}\}$, where $c_i(v)$ and $l_i(v)$ denote the minimum radius and maximum longitudinal slope of the ramp $v$, respectively.
\end{itemize}
\end{defn}


In general, $L_i(v)$, $c_i(v)$, and $l_i(v)$ can be any possible values.
However, to guarantee safety, there are some constraints by regulations during the construction of highways.
In this paper, according to China's regulations on the design of highway interchanges \cite{china-guidelines}, we have $L_i(v)\in \{3,4,5\}$, $c_i(v)\in \{\infty, 280,210,150,100,60,40,30\}$, and $l_i(v)\in \{1,2,3,4,5\}$.
The generation of concrete interchanges aims to cover as many value combinations of the three features as possible.
However, full combination sampling will result in an exponential increase in the number of interchanges.
For example, the topology graph of intersection $J_1$ (Fig.~\ref{fig:example_interchange_graph}) has four roads and four ramps. Full combination sampling will result in ${|L_i(v)|}^4 \times (|c_i(v)| \times |l_i(v)|)^4$ $=$ $3^4 \times (8 \times 5)^4$ $=$ $207,360,000$ interchanges.
Moreover, many generated interchanges by full combination sampling will show high similarity.
As a result, to preserve the geometrical diversity while minimizing duplication, we opt for 2-way combinations to balance the combinatorial coverage and the number of generated interchanges. Specifically, given a topology model $\mathbb{G}$ with $n$ roads and $m$ ramps, the number of parameters for sampling is $N_{param} = n \times |L(v)| + m \times (|c(v)|+|l(v)|)$. Thus, we can generate a set of interchange features
$\mathbb{F}=\{F_i: F_i=\langle \mathbb{G}^J, FR_i^J, Fr_i^J\}$, where $FR_i^J=(L_i(R_1), \ldots, L_i(R_n))$ and $Fr_i^J=(c_i(r_1), l_i(r_1)), \ldots, (c_i(r_m), l_i(r_m)))$,
such that the value combinations of any parameter pair in the $N_{param}$ parameters are covered.
In the sequel, given a topology model $\mathbb{G}$ and the feature sets $\mathbb{F}$,
we illustrate the process of generating concrete interchanges.


\begin{figure}
    \centering
    \subfigure[Bézier curve]{
        \begin{minipage}[b]{0.31\columnwidth}
            \centering
            \label{fig:bezier-curve-example}
            \includegraphics[width=1\textwidth]{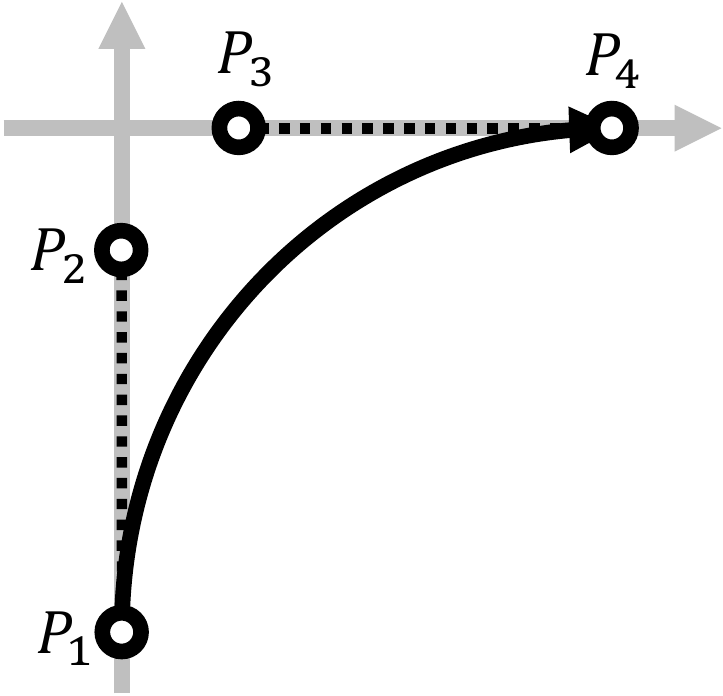}
        \end{minipage}}
    \subfigure[Bézier spline 1]{
        \begin{minipage}[b]{0.31\columnwidth}
            \centering
            \label{fig:bezier-spline-example}
            \includegraphics[width=1\textwidth]{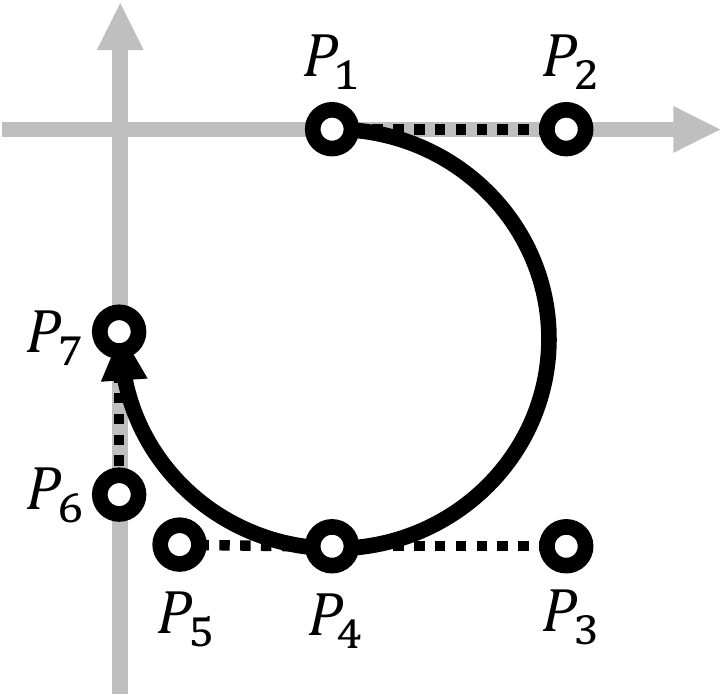}
        \end{minipage}}
    \subfigure[Bézier spline 2]{
        \begin{minipage}[b]{0.31\columnwidth}
            \centering
            \label{fig:bezier-spline-example-2}
            \includegraphics[width=1\textwidth]{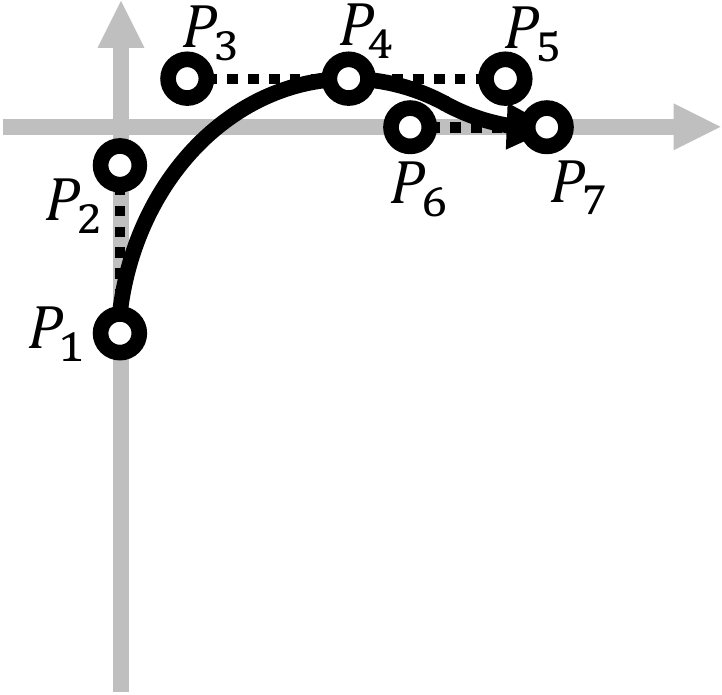}
        \end{minipage}}
    \caption{Ramps generated by Bézier curves (control points highlighted)}
    \label{fig:ramp-generation}
\end{figure}

To generate a concrete interchange, we need to generate the lane curve for each ramp based on the geometric parameters.
We apply B{\'e}zier curves~\cite{mortenson1999mathematics} to construct the lane curve in this paper.
B{\'e}zier curves are straightforward to construct and able to control the gradient at its start and end points, which is essential in smoothing the linkage between roads and ramps \cite{bezier_chen2013lane, bezier_kluck2021automatic, bezier_misro2019determining}
As shown in Fig.~\ref{fig:bezier-curve-example}, a B{\'e}zier curve $B(t)$ is constructed by four control points $P_1-P_4$ and can be formulated as $B(t)=(1-t)^3 P_0+3(1-t)^2 t P_1+3(1-t) t^2 P_2+t^3 P_3,\  t\in[0,1]$.
In this study, roads and ramps with small curvature or smooth shapes are built using quadratic B{\'e}zier curves (e.g., Fig.~\ref{fig:bezier-curve-example}), while ramps with large curvature or rough shapes are constructed by Bézier splines, each of which is a concatenation of  multiple Bézier curves (e.g., Fig.~\ref{fig:bezier-spline-example} and Fig.~\ref{fig:bezier-spline-example-2}).

We first generate the one-way roads. The one-way roads are primarily controlled by their endpoints, which are randomly positioned within their respective search areas.
Depending on the number of the one-way roads in the interchange, the search area for each endpoint of each one-way road is heuristically outlined within 2D free space, such that 1) the one-way roads are grouped into expressways, each of which contains at most two kinds of roads with opposite directions, and 2) the overall shape of the interchange mimics those of the real-world interchanges (e.g., T-shaped interchange for three roads and cross-shaped interchanges for four roads).

After the one-way roads are pinned, we generate the ramps connecting the roads based on the given topology model.
For each ramp, we apply differential evolution \cite{storn1997differential} to search for optimal locations of the control points, subject to the following constraints: 1) the ramp should start from the center of its predecessor (a road or another ramp) and end at the center of its successor (a road or another ramp); 2) the transition between the ramp and its predecessor (and successor) should be smooth; 3) the B{\'e}zier spline (if any) should be smooth; 4) the resulting curve should satisfy both the minimum radius and the maximum longitudinal slope requirements.
Algorithm \ref{alg:search_process} shows the detailed search process.

\begin{algorithm}[t]
    \small
    \caption{Search control points based on differential evolution.}
    \label{alg:search_process}
    \KwIn{A set of sampled ramp features of an interchange $\mathbb{F} = \{F_1, F_2, ..., F_K\}$}
    \KwResult{A set of Bézier curves of ramps $\mathbb{B} = \{B_1, B_2, ..., B_n\} $}
    Initialization: $\mathbb{B} = \emptyset$\;
    \For{$F_i \in \mathbb{F}$} {
        $success \gets false$\;
        retrieve the target radius  $r_\mathrm{target}$ and slope $s_\mathrm{target}$ from $F_i$\;
        Initialize the parameters randomly $param$\;
        \For{$count < count_\mathrm{max}$} {
            Generate the B{\'e}zier curve $B_i$ based on $param$\;\tcc{Constraints 1 and 2 are satisfied.}
            retrieve the curve $curv$, minimum radius $r_\mathrm{min}$, and maximum slope $s_\mathrm{max}$ from $B_i$\;
            $p_\mathrm{curv} \gets$ calculate\_variance($curv$)\;
            $p_\mathrm{radius} \gets (r_\mathrm{min} – r_\mathrm{target})^2$\;
            $p_\mathrm{slope} \gets (s_\mathrm{max} – s_\mathrm{target})^2$\;
            $penal \gets p_\mathrm{curv} + p_\mathrm{radius} + p_\mathrm{slope}$\;

            \If{$penal < penal_\mathrm{min}$}
            {
            \tcc{Constraints 3 and 4 are satisfied.}
                $\mathbb{B} = \mathbb{B} \cup \{B_i\}$\;
                $success \gets true$\;
                \textbf{break}\;
            }
            \Else
            {
                $param \gets$ differential\_evolution($param$, $penal$)
            }
            $count = count + 1$
        }
        \If{not $success$} {
            resampling\_parameters($i$)

            exit()
        }
    }

\end{algorithm}

\section{Experiments}

The experiments aim to answer the following research questions.

\noindent \textbf{RQ1}: What kinds of interchanges can \tool generate?

\noindent \textbf{RQ2}: What are the inter-class and intra-class differences of the generated interchanges?

\noindent \textbf{RQ3}: Can the generated interchange dataset be applied to evaluate an AV's algorithms?

We implement a prototype of \tool to generate a data set of interchanges in the OpenDRIVE format \cite{web_openDRIVE} for greater compatibility. We use Microsoft's PICT tool~\cite{github_microsoft_pict} to perform the 2-way combinatorial sampling and the Scipy\footnote{\url{https://docs.scipy.org/doc/scipy/reference/generated/scipy.optimize.differential_evolution.html}}
to optimize the control points' positions in the 3D space.

To answer RQ1, we conduct qualitative analysis by visually comparing the interchanges from the same and different topology classes.
To answer RQ2, we perform a comprehensive throughput test with SUMO's in-built traffic flow control algorithm.
Finally, to demonstrate the applicability of the generated dataset, we test the fuel-optimization trajectory tracking algorithm deployed to Alibaba's autonomous trucks on the generated interchanges.
The complete experimental results can be found at \website.

\subsection{Examples of Generated Interchanges}


\setlength{\tabcolsep}{2.2pt}
\begin{table}
\caption{The number of interchange samples per topology class}
\vspace{-10pt}
\label{tab:samples-per-class}
\begin{tabular}{cccccccccccc}
\hline
\begin{tabular}[c]{@{}c@{}}Topology\\ Class\end{tabular} & $C_1$ & $C_2$ & $C_3$ & $C_4$ & $C_5$ & $C_6$ & $C_7$ & $C_8$ & $C_9$ & $C_{10}$ & $C_{11}$ \\ \hline
\begin{tabular}[c]{@{}c@{}}Interchange\\ Samples\end{tabular} & 74 & \textbf{77} & \textbf{77} & \textbf{77} & 73 & \textbf{77} & 61 & 61 & 61 & 73 & 61 \\ \hline
\multicolumn{1}{l}{} & \multicolumn{1}{l}{} & \multicolumn{1}{l}{} & \multicolumn{1}{l}{} & \multicolumn{1}{l}{} & \multicolumn{1}{l}{} & \multicolumn{1}{l}{} & \multicolumn{1}{l}{} & \multicolumn{1}{l}{} & \multicolumn{1}{l}{} & \multicolumn{1}{l}{} & \multicolumn{1}{l}{} \\ \hline
\begin{tabular}[c]{@{}c@{}}Topology\\ Class\end{tabular} & $C_{12}$ & $C_{13}$ & $C_{14}$ & $C_{15}$ & $C_{16}$ & $C_{17}$ & $C_{18}$ & $C_{19}$ & $C_{20}$ & $C_{21}$ & Total \\ \hline
\begin{tabular}[c]{@{}c@{}}Interchange\\ Samples\end{tabular} & 61 & \textbf{77} & 61 & 61 & 61 & 73 & 73 & \textbf{50} & \textbf{77} & \textbf{77} & 1443 \\ \hline
\end{tabular}
\end{table}

In our example, we search for the interchanges in Hangzhou using the keyword ``highway interchange'' on the GaoDe map, resulting in 39 interchanges in 2D space.
We then apply \tool to normalize the interchanges' locations, construct their topology models, and classify them into 21 topology-equivalency classes. For each class, \tool uses 2-way combinatorial sampling and differential evolution to generate 1443 interchange samples.
The detailed statistic data is given in Table.~\ref{tab:samples-per-class}.

\begin{figure}
    \centering
    \subfigure[]{
        \begin{minipage}[b]{0.3\columnwidth}
            \centering
            \label{fig:zijingang-interchange}
            \includegraphics[width=1\textwidth]{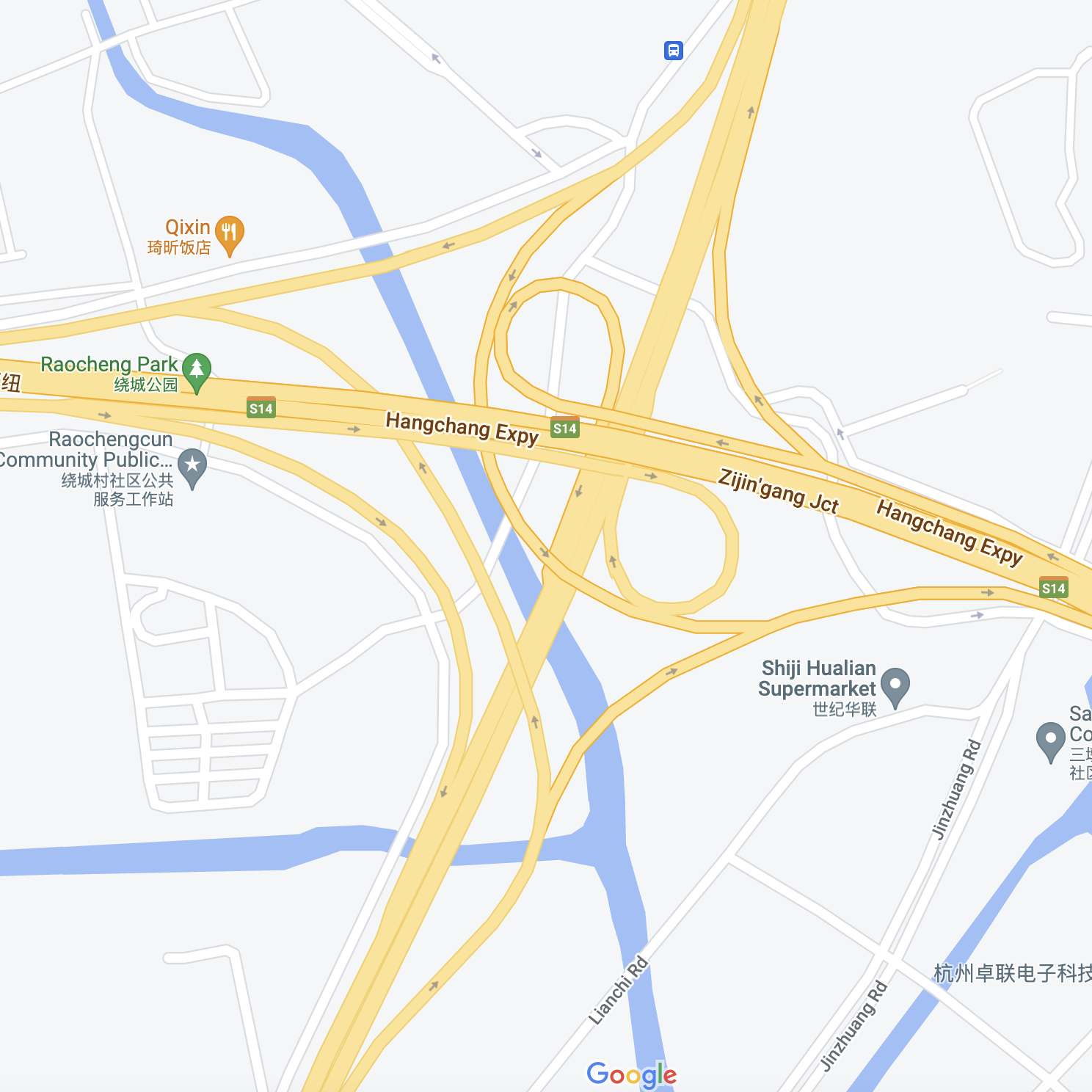}
        \end{minipage}}
    \subfigure[]{
        \begin{minipage}[b]{0.3\columnwidth}
            \centering
            \label{fig:zijingang-diagram}
            \includegraphics[width=1\textwidth]{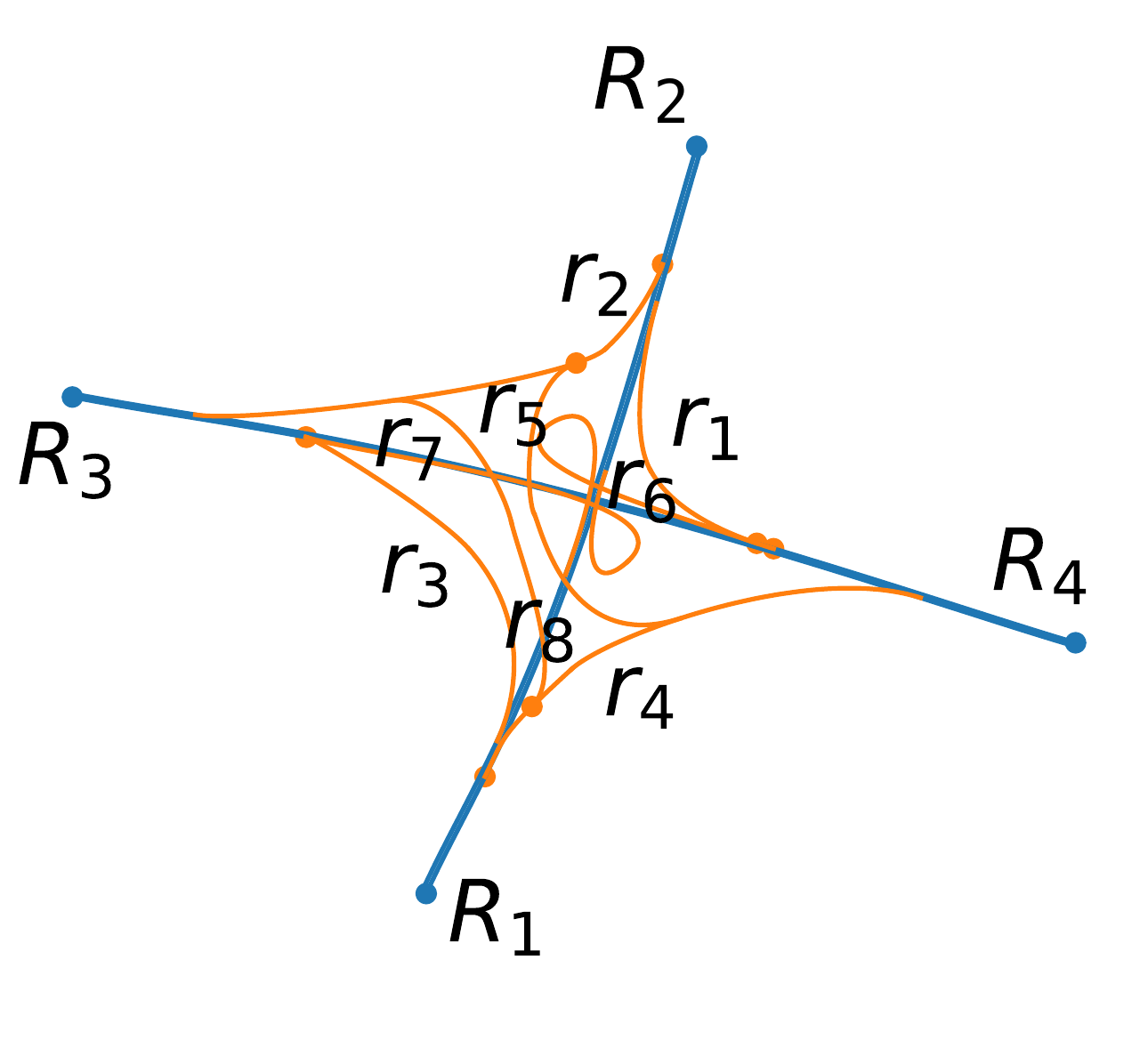}
        \end{minipage}}
    \subfigure[]{
        \begin{minipage}[b]{0.3\columnwidth}
            \centering
            \label{fig:zijingang-topology-graph}
            \includegraphics[width=1\textwidth]{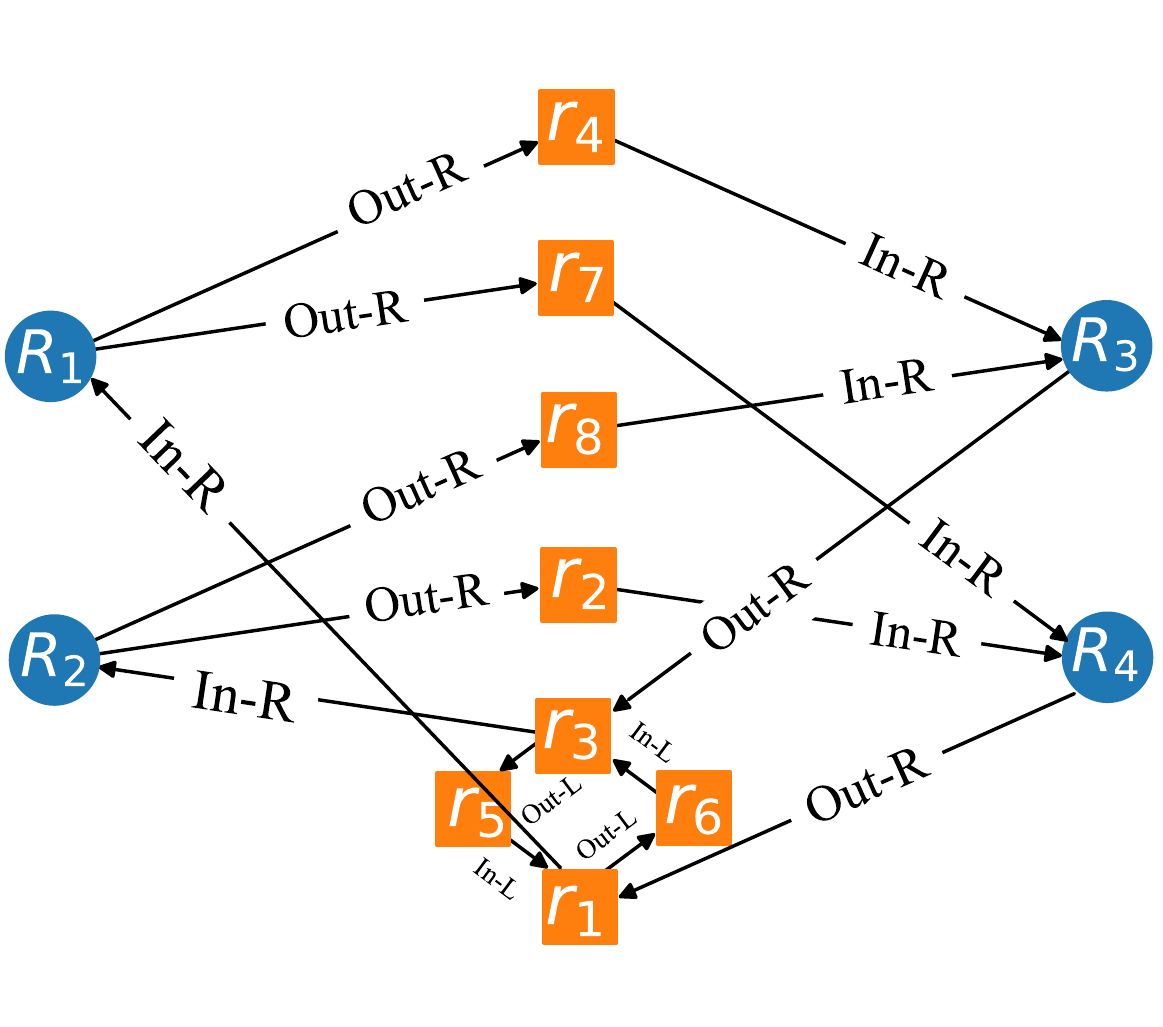}
        \end{minipage}}
    \subfigure[]{
        \begin{minipage}[b]{0.3\columnwidth}
            \centering
            \label{fig:sample-1-from-C6}
            \includegraphics[width=1\textwidth]{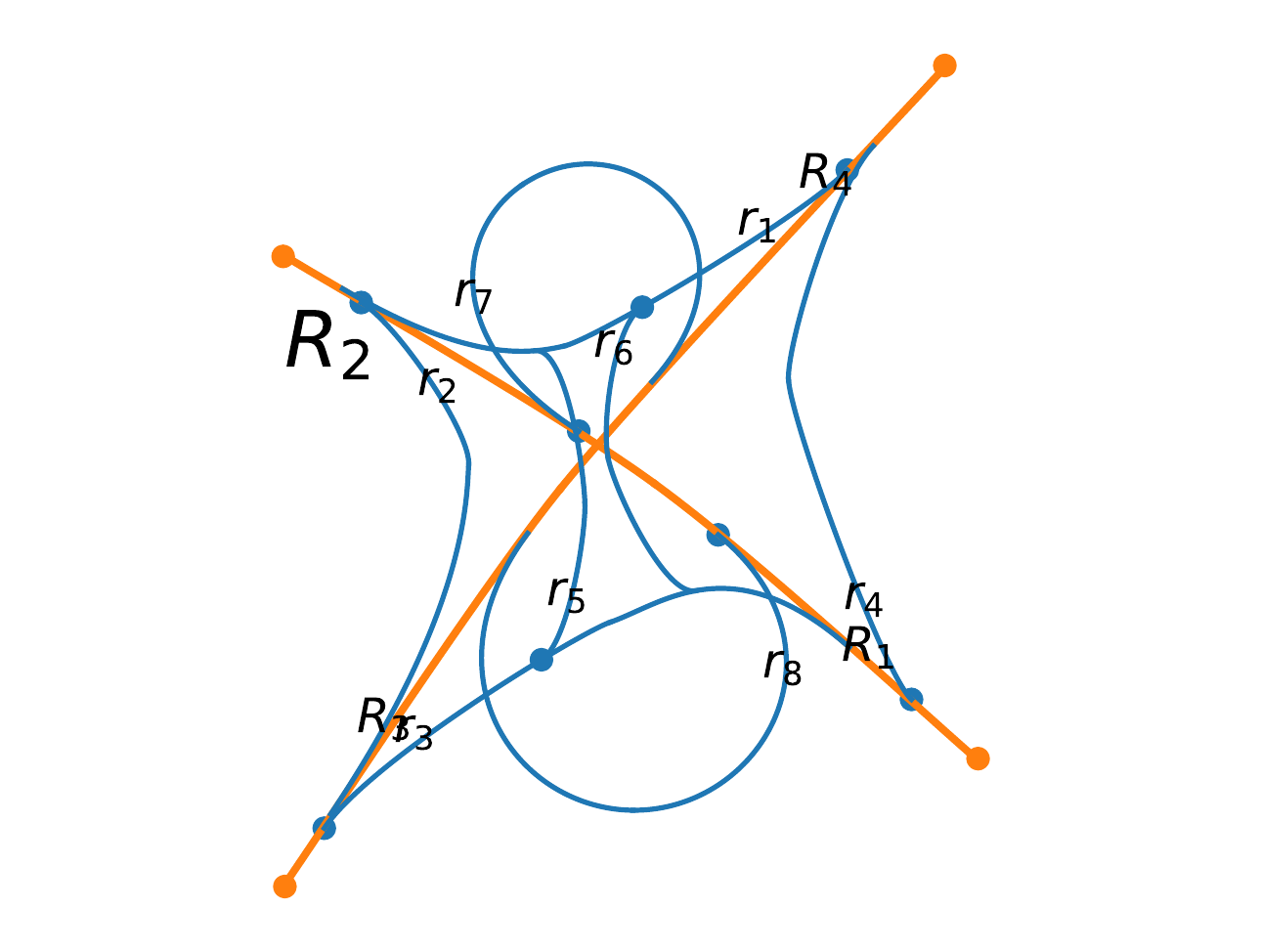}
        \end{minipage}}
    \subfigure[]{
        \begin{minipage}[b]{0.3\columnwidth}
            \centering
            \label{fig:sample-2-from-C6}
            \includegraphics[width=1\textwidth]{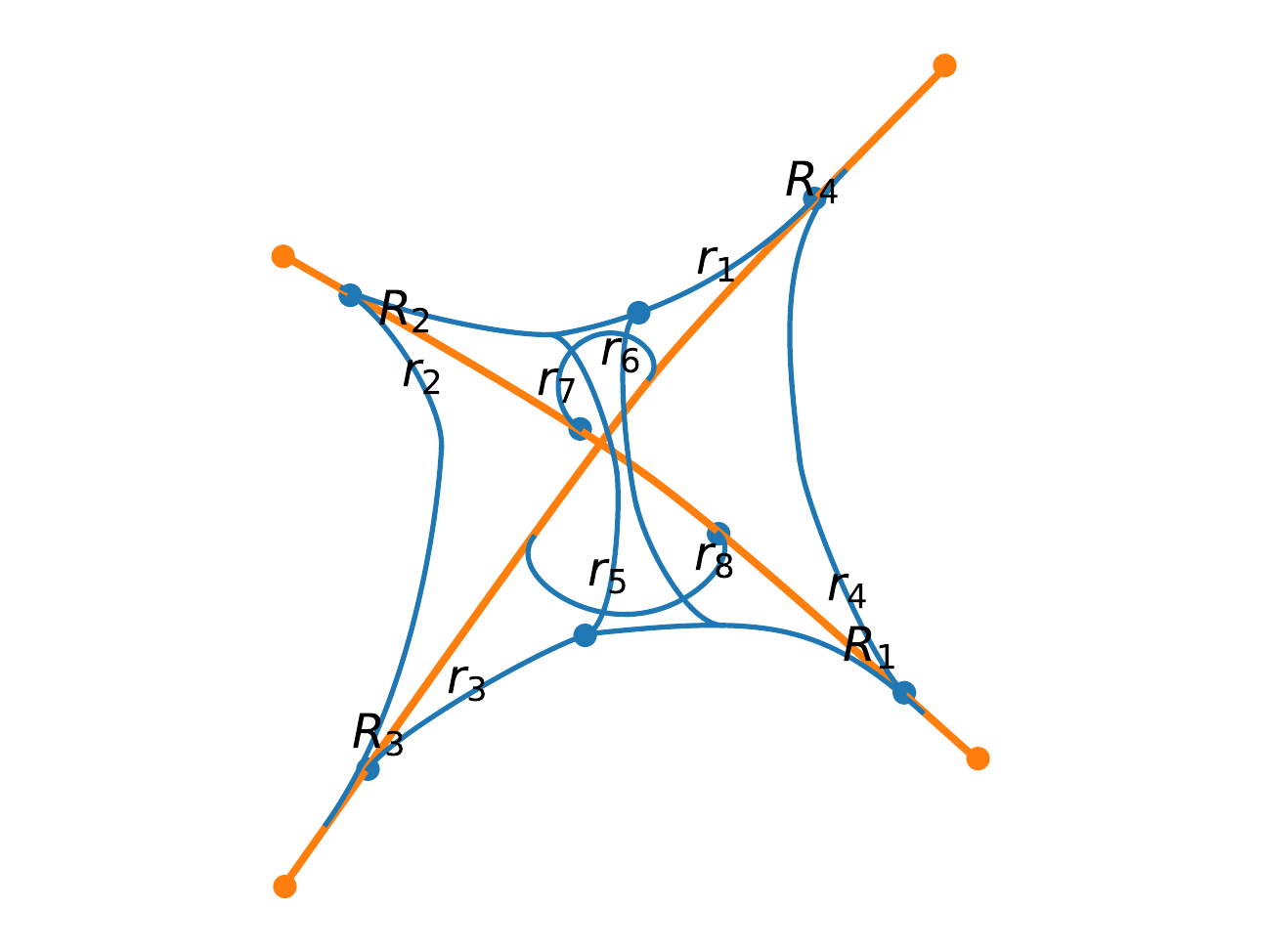}
        \end{minipage}}
    \subfigure[]{
        \begin{minipage}[b]{0.3\columnwidth}
            \centering
            \label{fig:sample-3-from-C6}
            \includegraphics[width=1\textwidth]{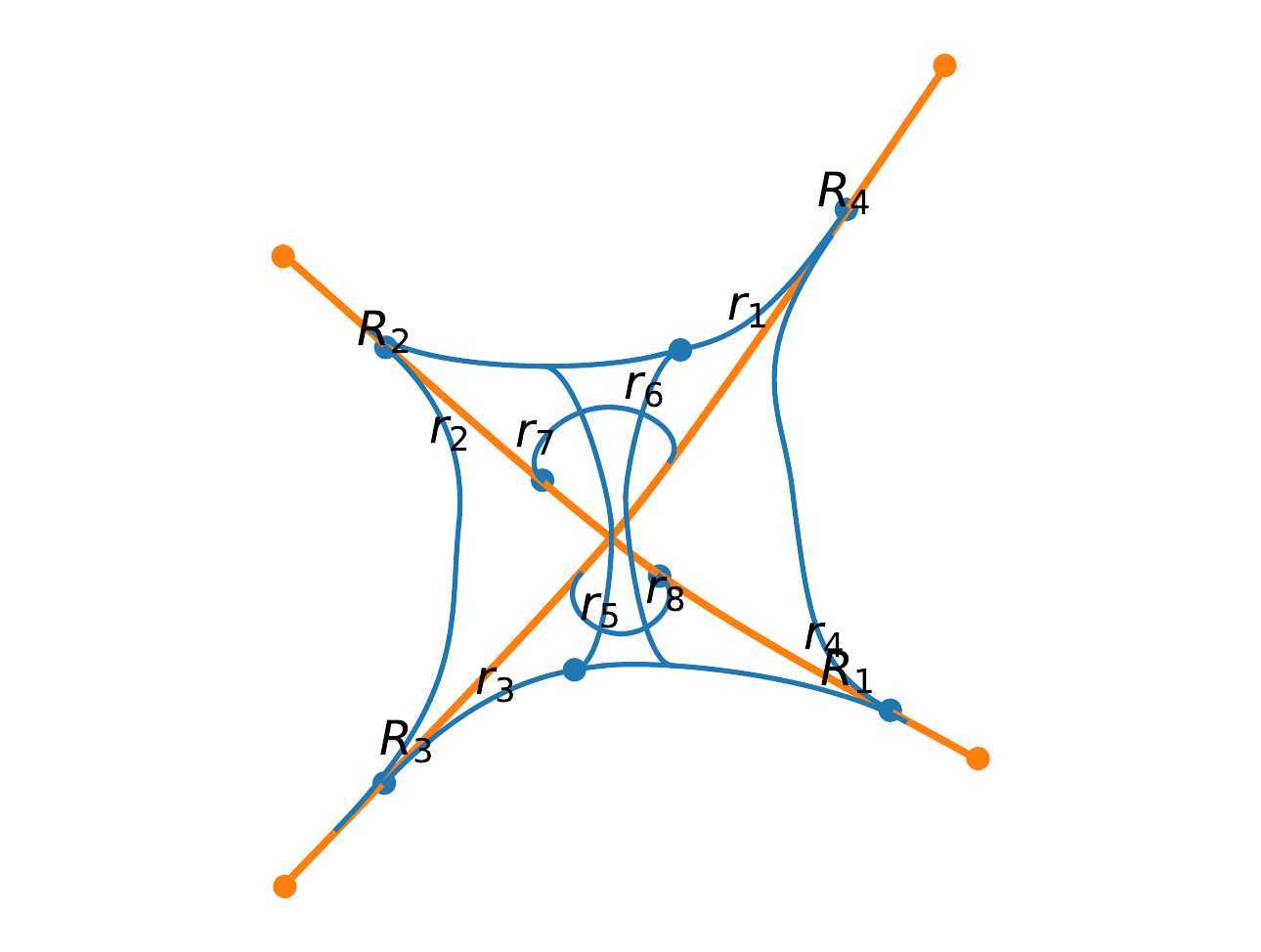}
        \end{minipage}}
      \vspace{-5pt}
    \caption{(a) The ZiJinGang Interchange $J_Z$; (b) $J_Z$'s diagram, (c) $J_Z$'s topology graph; (d-f) the diagrams of three generated samples from the $J_Z$'s topology class.
    }
    \label{fig:zijingang-example}
\end{figure}

Fig. \ref{fig:zijingang-example} shows an example of the  ZiJinGang Interchange (Fig.~\ref{fig:zijingang-interchange}), denoted as $J_Z$.
Based on the data generated from the real-world map, the diagram of interchange $J_Z$ is given in Fig.~\ref{fig:zijingang-diagram}, which contains four one-way roads and eight ramps.
The topology model $\mathbb{G}_{J_Z}$ is presented in Fig.~\ref{fig:zijingang-topology-graph}, which is classified into class $C_6$ out of the  21 classes.
The diagrams of three interchanges out of 72 samples from $C_6$ are presented in Figs.~\ref{fig:sample-1-from-C6}--\ref{fig:sample-3-from-C6}.
It can be seen that after the geometric parameter sampling and control points optimization, \tool generates interchanges of varying curvatures and slopes (which depend on the traveling distance of each ramp, assuming the height between the expressways is the same) from the same topology class.
The dataset of the generated interchanges is given at \website.


\subsection{Evaluation on Traffic Flow Control Algorithms}
\begin{figure}
  \centering
   \includegraphics[width=\columnwidth]{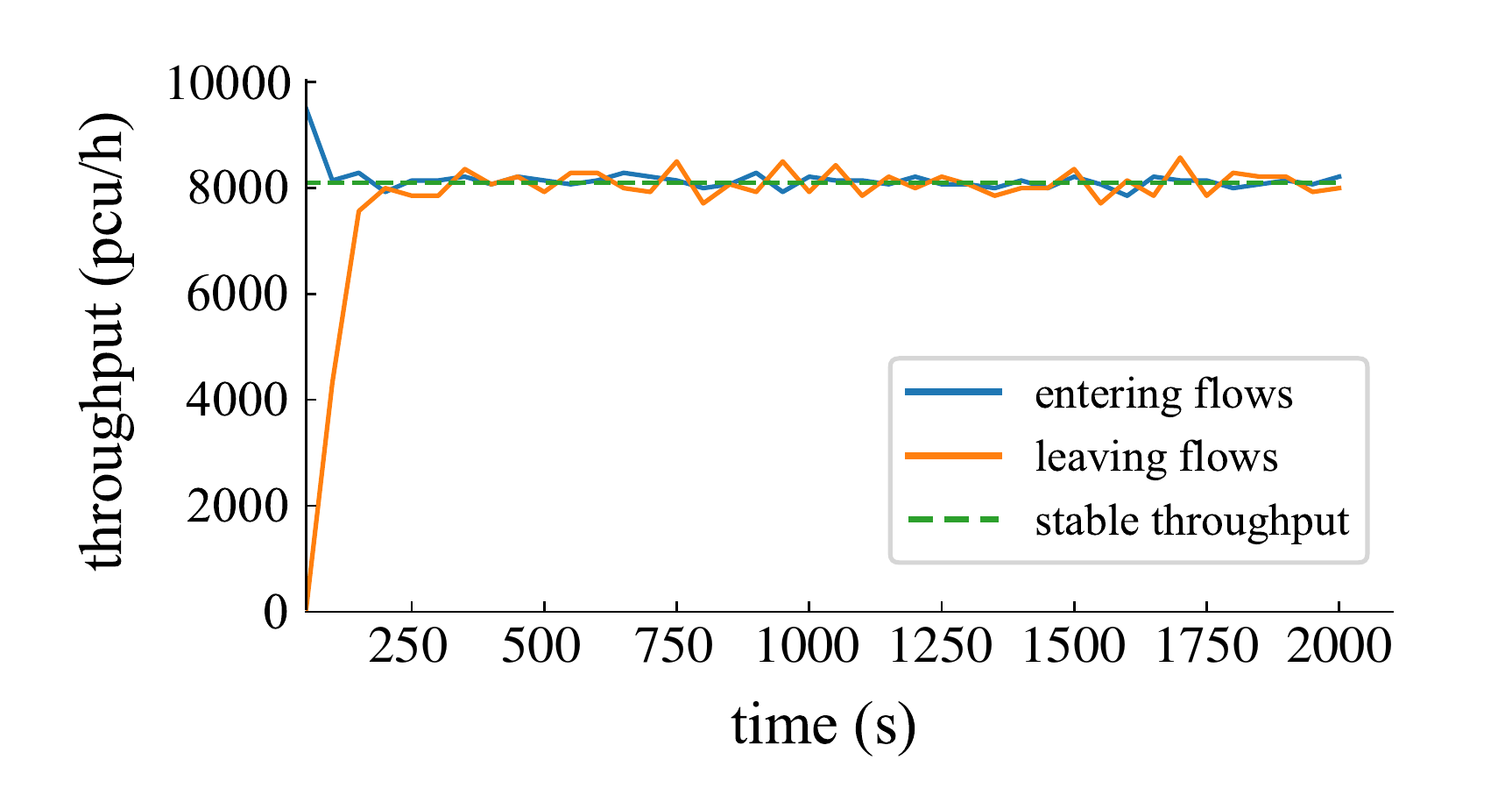}
   \vspace{-20pt}
   \caption{Throughput of the entering and leaving vehicles  in an interchange of $C_3$.}
   \label{fig:changes_of_vehicle_numbers}
\end{figure}

In this section, we run SUMO's built-in traffic flow control algorithm with the generated dataset and use throughput to illustrate the diversity of the interchanges among different classes and in the same class.
We use SUMO (version 1.13.0) to inject traffic flows, with a predefined density, into the interchanges.
For each interchange, we assign a traffic flow for every entrance-exit pair and run the simulation for 2000 seconds.
The traffic flow of the interchange reaches a stable state if the entering and the leaving flows reach a balance.
For example, Fig. \ref{fig:changes_of_vehicle_numbers} shows the throughput of the entering and leaving vehicles in an interchange.
From the figure, we can find that after 250 seconds, the traffic flow in the interchange is stable, and the stable throughput is 8087.4 pcu/h (passenger car unit per hour).

\begin{figure}
  \centering
   \includegraphics[width=\columnwidth]{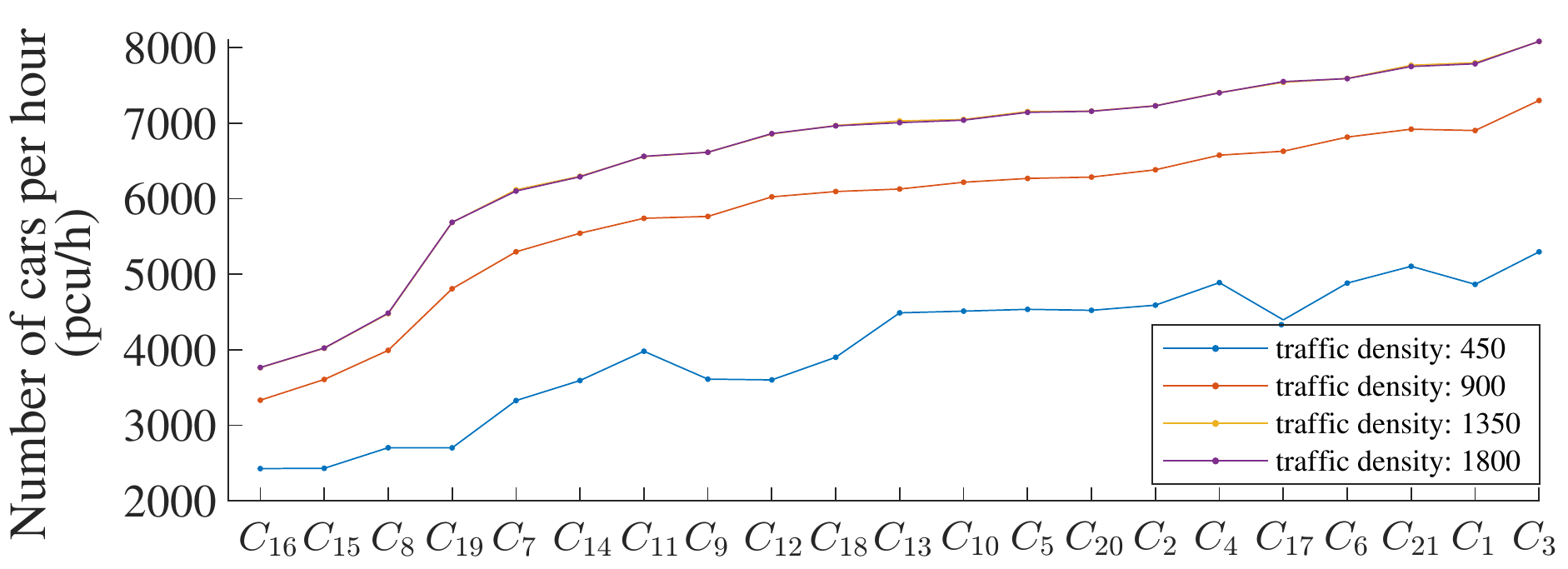}
   \caption{Average throughput of the 21 interchange classes, sorted by the throughput value at highest density.}
   \label{fig:class-average-throughput}
\end{figure}

\begin{figure}
    \centering
    \subfigure[450]{
        \begin{minipage}[b]{0.22\columnwidth}
            \centering
            \label{fig:class-throughput-t-test-450}
            \includegraphics[width=1\textwidth]{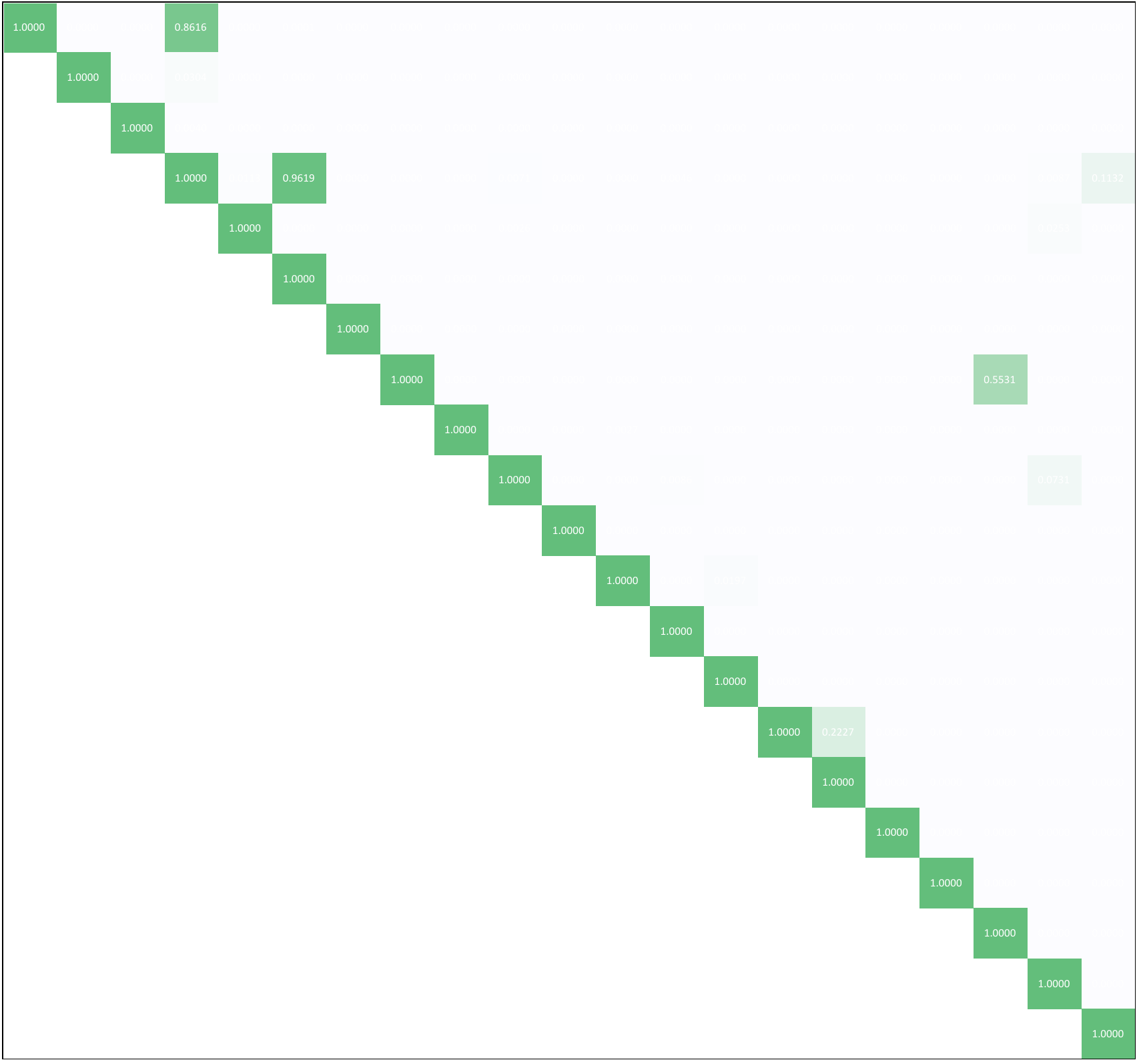}
        \end{minipage}}
    \subfigure[900]{
        \begin{minipage}[b]{0.22\columnwidth}
            \centering
            \label{fig:class-throughput-t-test-900}
            \includegraphics[width=1\textwidth]{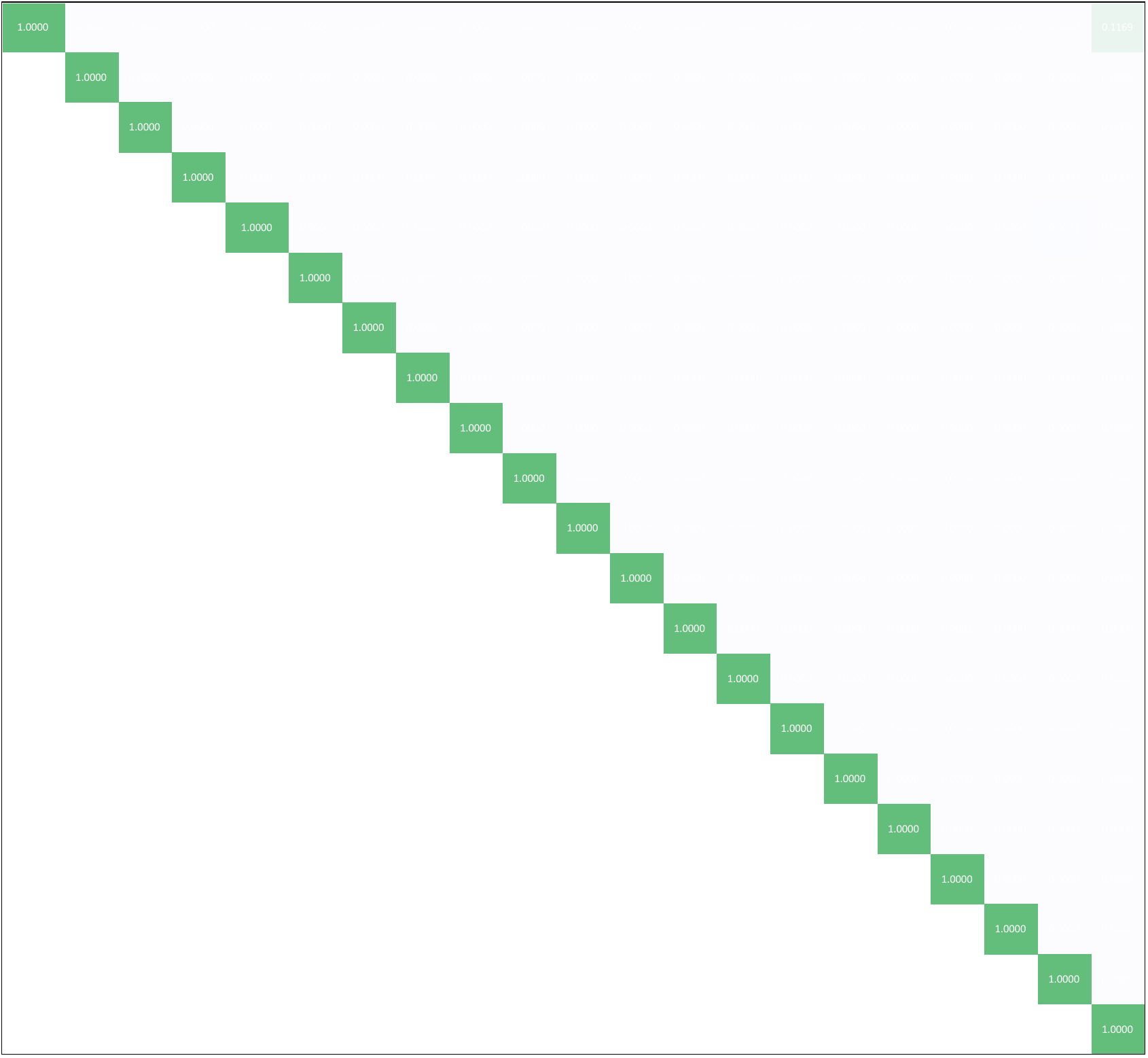}
        \end{minipage}}
    \subfigure[1350]{
        \begin{minipage}[b]{0.22\columnwidth}
            \centering
            \label{fig:class-throughput-t-test-1350}
            \includegraphics[width=1\textwidth]{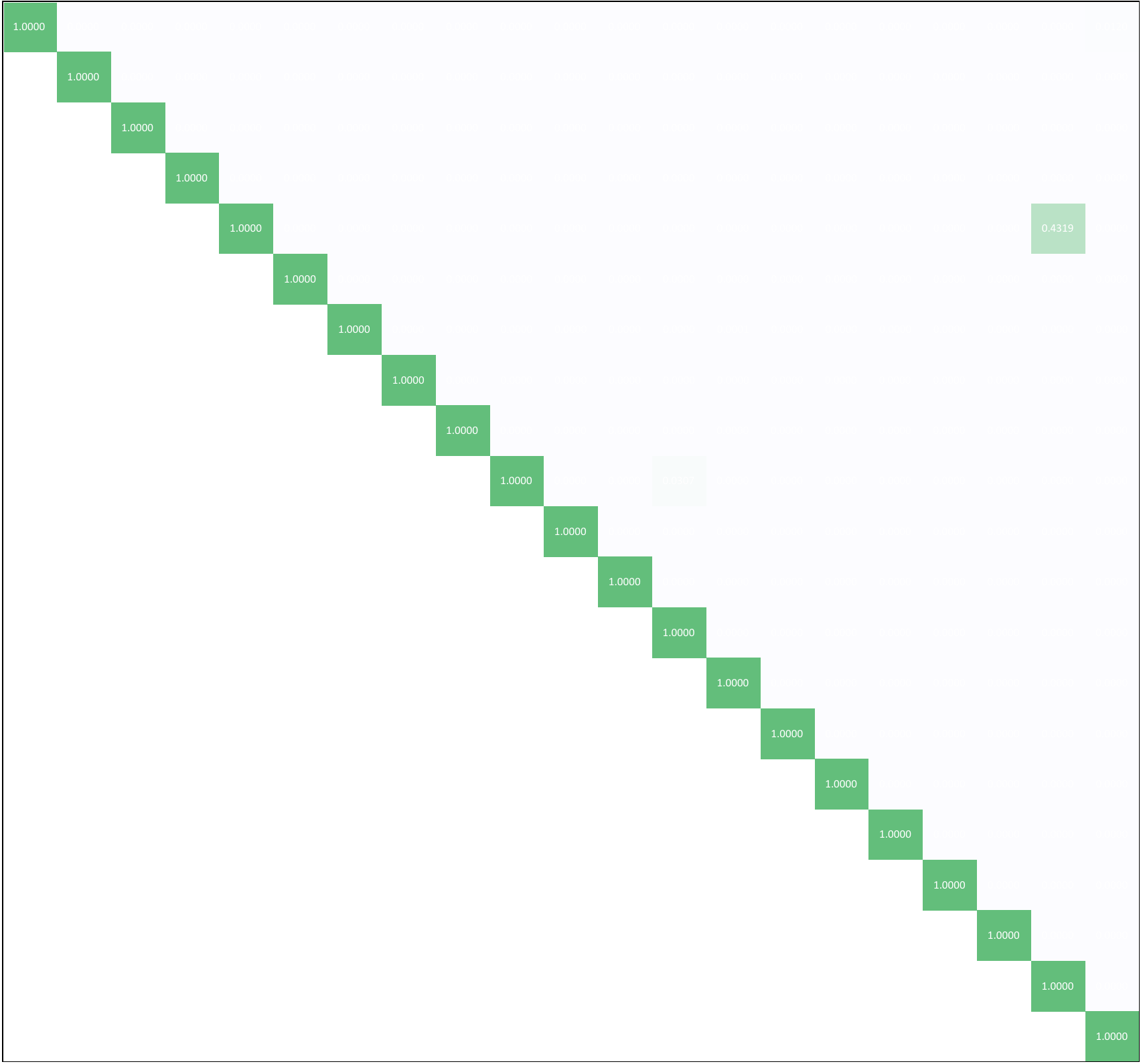}
        \end{minipage}}
    \subfigure[1800]{
        \begin{minipage}[b]{0.22\columnwidth}
            \centering
            \label{fig:class-throughput-t-test-1800}
            \includegraphics[width=1\textwidth]{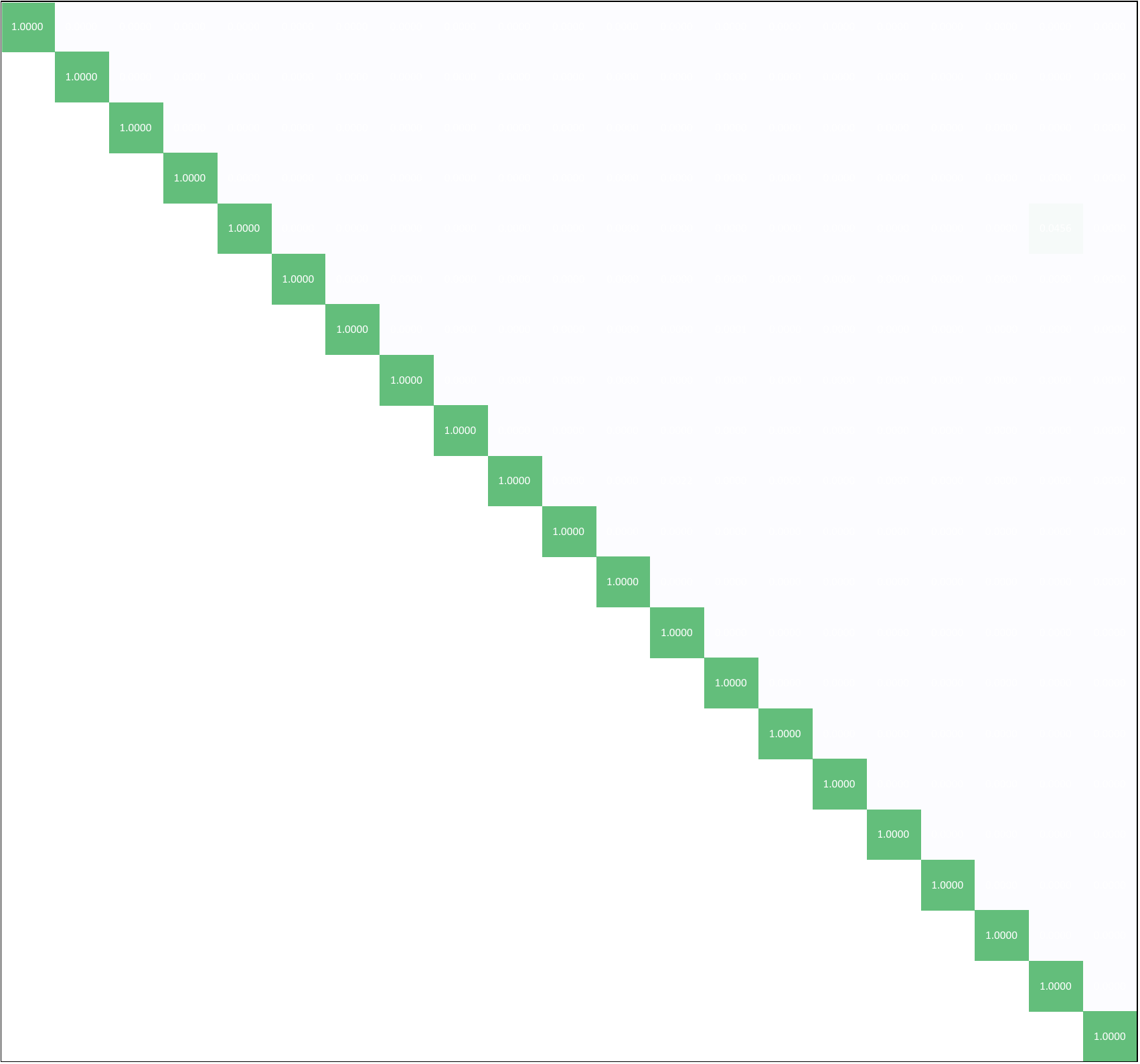}
        \end{minipage}}
    \caption{T-test for the throughputs among the 21 topology classes at different traffic densities.
    The X and Y axes are the topology classes from $C_{1}$ to $C_{21}$, and the color changes from white to full green as the t-value increases from 0 to 1.0.}
    \label{fig:inter-class-throughput-t-tests}
\end{figure}

We first compare the average throughput across the 21 interchange classes.
For each class, we randomly select 20 interchange samples and compute their average throughput.
The results are given in Fig.~\ref{fig:class-average-throughput}, where each line shows the average throughput of the 21 classes at a given traffic density.
It can be seen that: 1) The throughput increases with the traffic density in general; 2) the throughput reaches the maximum capacity when the traffic density is larger than 1350~pcu/h, resulting in the almost overlapping lines for 1350~pcu/h and 1800~pcu/h; 3) the interchanges from different topology classes manifest different throughput profile. Note that compared to the rest cross-shaped interchanges, $C_{16}$, $C_{15}$ and $C_8$ are T-shaped interchanges with less roads, so they show smaller throughputs.
$C_3$ is a typical fully-connected ``Cloverleaf'' interchange, so it has the highest throughput.
To evaluate the statistical significance of the differences in the average throughputs of the topology classes, we conduct t-test.
The results are given in Fig.~\ref{fig:inter-class-throughput-t-tests}.
We find that given the significance level $\alpha=0.05$, some interchange classes are not statistically significant at a low traffic density (e.g., 450),  while they are statistically significant when the traffic density increases.
It is because some interchanges do not reach the maximal capacities at a low traffic density. Thus, they are not statistically significant at the throughput.


\begin{figure}
    \centering
    \subfigure[450]{
        \begin{minipage}[b]{0.22\columnwidth}
            \centering
            \label{fig:intra-throughput-t-test-450}
            \includegraphics[width=1\textwidth]{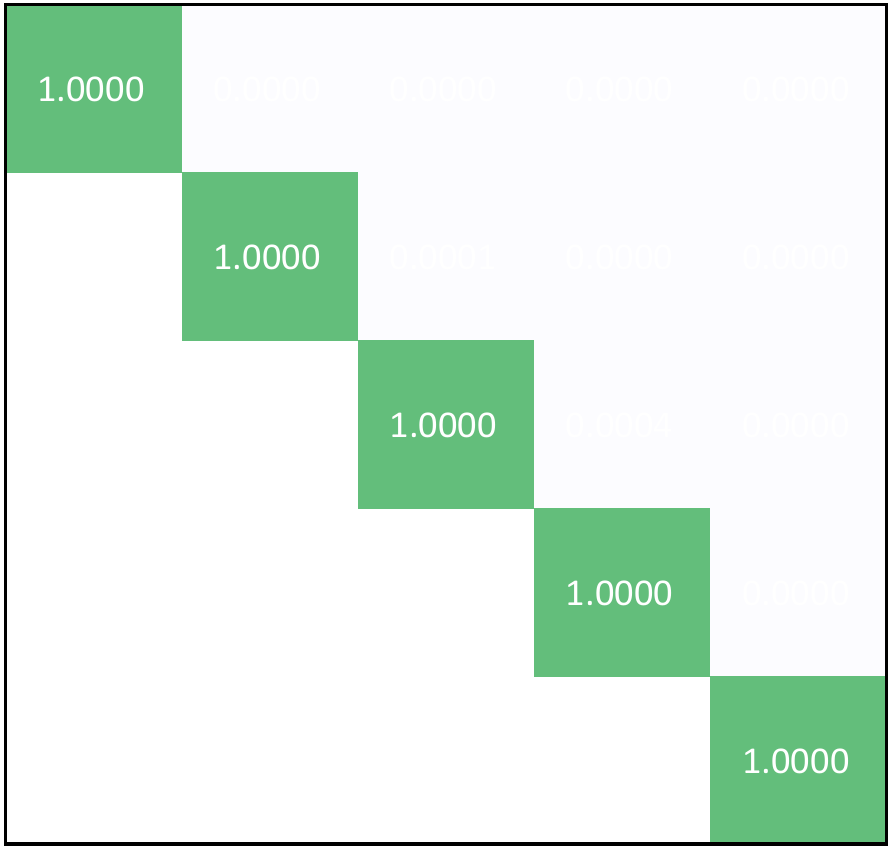}
        \end{minipage}}
    \subfigure[900]{
        \begin{minipage}[b]{0.22\columnwidth}
            \centering
            \label{fig:intra-throughput-t-test-900}
            \includegraphics[width=1\textwidth]{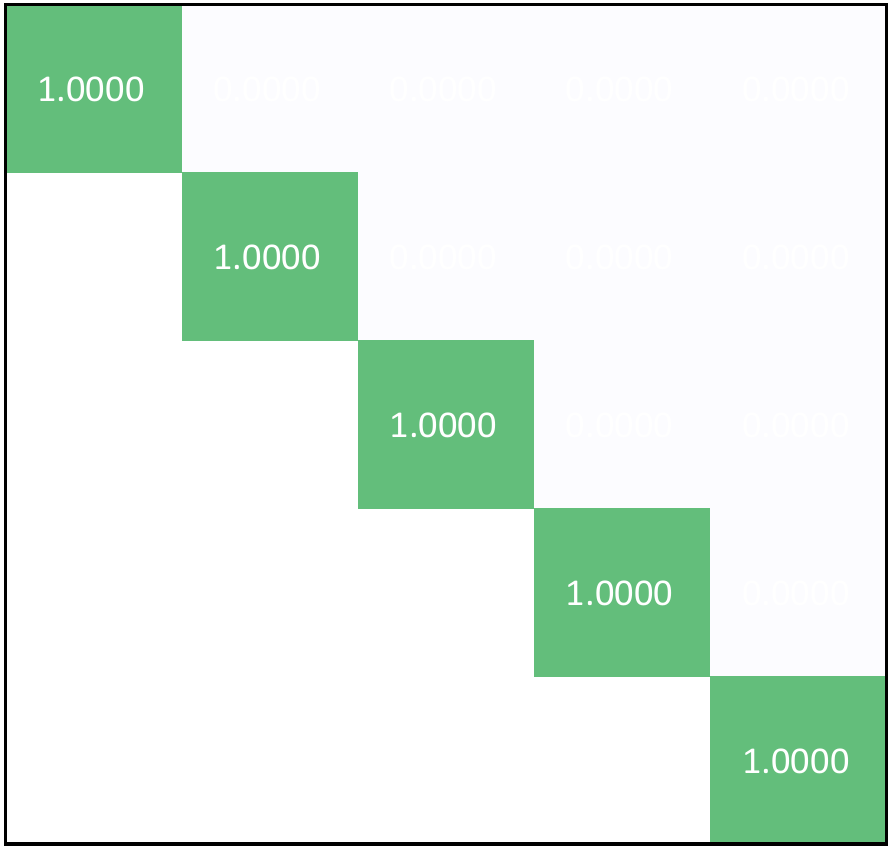}
        \end{minipage}}
    \subfigure[1350]{
        \begin{minipage}[b]{0.22\columnwidth}
            \centering
            \label{fig:intra-throughput-t-test-1350}
            \includegraphics[width=1\textwidth]{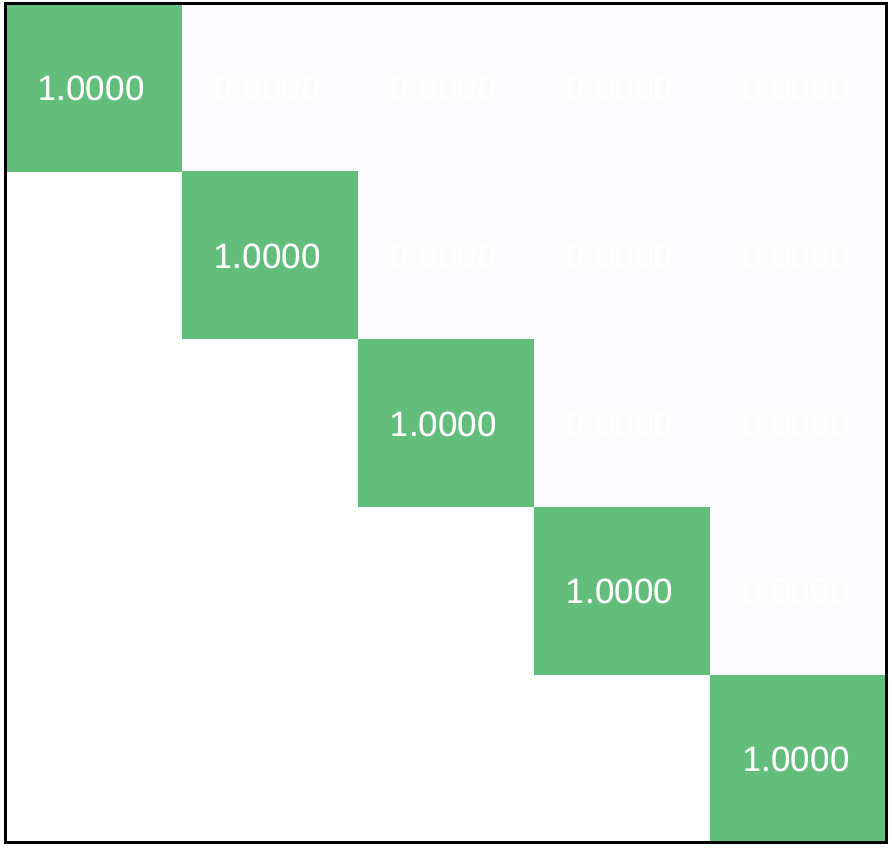}
        \end{minipage}}
    \subfigure[1800]{
        \begin{minipage}[b]{0.22\columnwidth}
            \centering
            \label{fig:intra-throughput-t-test-1800}
            \includegraphics[width=1\textwidth]{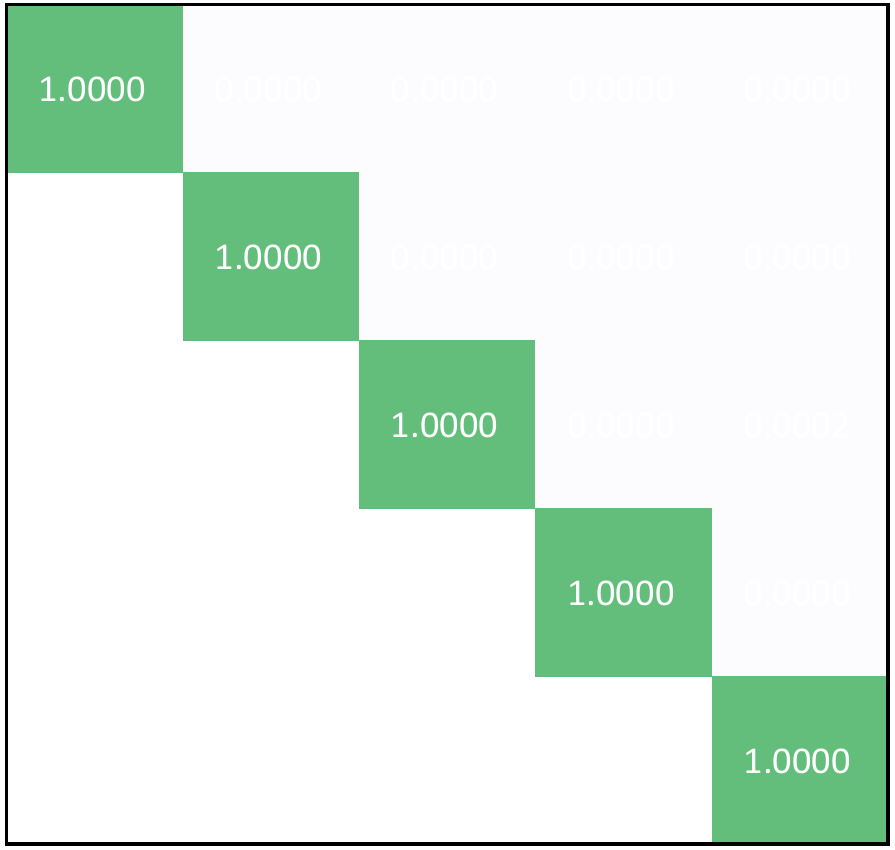}
        \end{minipage}}
    \caption{T-test for the throughputs among the five interchange samples in $C_6$.}
    \label{fig:intra-class-throughput-t-tests}
\end{figure}

Next, we show the difference in the intra-class throughput by comparing the interchanges within the same class.
Due to limited space, we only present the results of 5 interchange samples from $C_6$.
The throughput of each interchange and the results of the statistical significance testing are given in Table.~\ref{tab:intra-class-throughput-comparison} and Fig.~\ref{fig:intra-class-throughput-t-tests}, respectively.
The results indicate that the throughputs of the five interchanges are statistically significant.
It validates that with the same topology, the interchange geometrical features defined by \tool have an impact on the throughput.


\begin{table}
\centering
\vspace{-10pt}
\caption{Average throughput of five random interchanges in $C_6$}
\begin{tabular}{c|cccc}
\toprule
\multirow{2}{*}{Interchange} & \multicolumn{4}{c}{Car num per hour (pcu/h)} \\
                       & 450        & 900       & 1350      & 1800      \\
\midrule
$C_6$-1                  & 4703.58    & 6752.80    & 7622.66   & 7627.63   \\
$C_6$-2                  & 4768.72    & 6814.60    & 7592.32   & 7588.11   \\
$C_6$-3                  & 4820.40    & 6769.69    & 7619.85   & 7618.02   \\
$C_6$-4                  & 4893.82    & 6760.80    & 7576.71   & 7614.09   \\
$C_6$-5                  & 4917.42    & 6889.74    & 7652.34   & 7653.96   \\
\bottomrule
\end{tabular}
\label{tab:intra-class-throughput-comparison}
\end{table}

Finally, we show the difference in the throughputs of the 21 interchange class and the 5 samples in the same class.
Table \ref{tab:std} shows their standard deviations at different traffic densities.
The results show that the throughputs of the interchanges among different classes are  more spread out than those in the same class.
It also demonstrates the effectiveness of \tool.

\begin{table}[]
\centering
\vspace{-10pt}
\caption{Standard Deviations of the Inter-Class and Intra-Class Throughputs}
\label{tab:std}
\begin{tabular}{c|cccc}
\hline
&  \multicolumn{4}{c}{Car num per hour (pcu/h)}\\
& 450    & 900     & 1350    & 1800    \\ \hline
inter-class & 873.63 & 1062.44 & 1184.72 & 1181.98 \\
intra-class & 35.46  & 93.21   & 97.83   & 102.87  \\ \hline
\end{tabular}
\end{table}

\subsection{Database Application on Autonomous Trucks' Fuel Consumption Evaluation}

For heavy vehicles such as autonomous trucks,  the topology and geometry of highway interchanges are nontrivial.
Hence,  evaluating the developed optimal control systems for heavy vehicles requires an interchange dataset of varying topological and geometrical profiles.
In this section, we demonstrate an application of our interchange dataset in evaluating the fuel consumption of the trajectory tracking algorithm deployed to Alibaba’s autonomous trucks.
We use TruckSim \cite{truckSim} to build the high-fidelity truck model, whose trajectory is generated by a lattice planner \cite{mcnaughton2011motion} and followed by a LQR controller \cite{LOPEZ20112230}.

\begin{table}
\centering
\vspace{-10pt}
\caption{Fuel consumption on individual ramps (ml/KM)}
\label{tab:ramp-fuel-consumption}
\begin{tabular}{c|cccc|cccc}
\hline
\multirow{2}{*}{\begin{tabular}[c]{@{}c@{}}Min\\ Radius\end{tabular}} & \multicolumn{4}{c|}{Up-ramp Slope (\%)} & \multicolumn{4}{c}{Down-ramp Slope (\%)} \\
 & 1 & 2 & 3 & 4 & 1 & 2 & 3 & 4 \\ \hline
30 & 366.33 & 384.93 & 406.23 & 431.40 & 321.21 & 314.31 & 312.81 & 311.67 \\
40 & 370.53 & 389.87 & 413.35 & 440.76 & 326.71 & 317.72 & 315.99 & 314.87 \\
60 & 377.79 & 398.33 & 423.92 & 453.54 & 334.96 & 323.71 & 321.61 & 320.52 \\
100 & 398.83 & 422.71 & 452.07 & - & 366.09 & 349.57 & 346.58 & - \\
150 & 447.65 & 475.10 & 508.42 & - & 441.3 & 433.94 & 430.31 & - \\
210 & 547.71 & 578.67 & 616.31 & - & 510.32 & 540.83 & 559.45 & - \\
280 & 702.46 & 737.32 & 779.68 & - & 637.86 & 630.62 & 619.12 & - \\ \hline
\end{tabular}
\end{table}

We first evaluate the fuel consumption per kilometer of the tracking algorithm at different ramp radii and slopes in both up-ramp and down-ramp scenarios.
Table.~\ref{tab:ramp-fuel-consumption} shows the fuel consumption at a preferred speed of $10m/s$.
It can be seen that, in both up-ramp and down-ramp scenarios, the slope and radius have a significant impact on fuel consumption (Table.~\ref{tab:ramp-fuel-consumption}). The fuel consumption increases with the increase of the ramp's minimum radius in both scenarios. On the other hand, as the slope increases, the truck consumes more fuels driving up-ramp and saves more fuel driving down-ramp.


\begin{figure}
  \centering
   \includegraphics[width=\columnwidth]{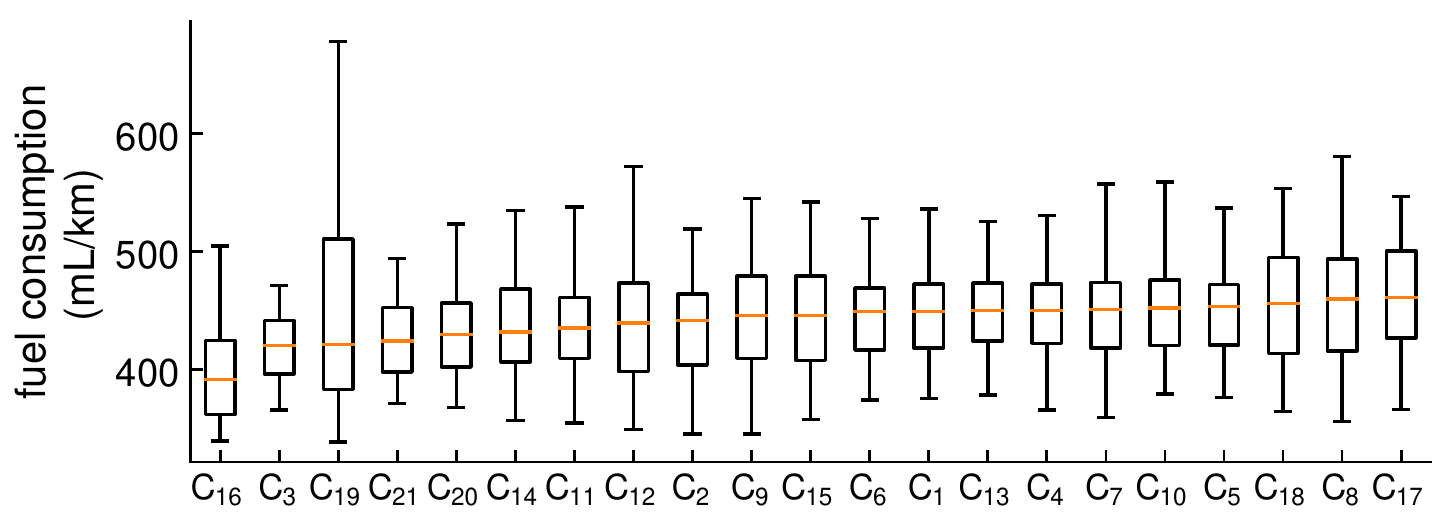}
   \vspace{-20pt}
   \caption{The distribution of fuel consumption for the 21 interchange classes, sorted by the average consumption.}
   \label{fig:inter-class-fuel-comparison}
\end{figure}

Next, based on the fuel consumption results on individual ramps, we estimate the average fuel consumption of each interchange in the dataset by averaging the fuel consumption of all its ramps.
The results are given in Fig.~\ref{fig:inter-class-fuel-comparison}.
From the results, we can find different interchange classes show different patterns of fuel consumption.
It demonstrates the necessity and usability of our dataset in evaluating the performance of AVs.


\section{Conclusions}
This paper proposes a model-driven method to generate diverse highway interchanges for AV testing.
First, the topology of an interchange is modeled as a labeled digraph.
Then \tool extracts topology models from real-world interchanges and classifies them into different classes.
For each topology class, \tool generates a set of concrete interchanges by sampling the geometrical features of the roads and ramps.
Experimental results show the diversity and applicability of the dataset generated by \tool.
In the future, we will generate more interchanges according to the real-world interchanges of different cities and evaluate different algorithms and AV systems on the dataset.


\bibliographystyle{IEEEtranS}
\bibliography{references}

\end{document}